\documentclass[11pt]{article}

\usepackage[preprint]{acl}

\usepackage{times}
\usepackage{latexsym}
\usepackage[T1]{fontenc}
\usepackage[utf8]{inputenc}
\usepackage{microtype}
\usepackage{inconsolata}
\usepackage{graphicx}

\usepackage{amsmath, amssymb, amsthm}
\usepackage{mathtools}
\usepackage{comment}
\usepackage{bm}
\usepackage{booktabs}
\usepackage{hyperref}
\usepackage{xcolor}
\usepackage{enumitem}

\newtheorem{lemma}{Lemma}
\newtheorem{theorem}{Theorem}

\newtheorem{assumption}{Assumption}

\DeclareMathOperator{\AUROC}{AUROC}
\DeclareMathOperator{\SNR}{SNR}
\DeclareMathOperator{\MC}{MCS}
\DeclareMathOperator{\Cov}{Cov}

\newcommand{\R}{\mathbb{R}}
\newcommand{\E}{\mathbb{E}}
\newcommand{\Sigt}{\Sigma_{\mathrm{tot}}}
\newcommand{\Sigp}{\Sigma_{\mathrm{pool}}}
\newcommand{\wid}{w_{\mathrm{id}}}
\newcommand{\wood}{w_{\mathrm{ood}}}
\newcommand{\zmax}{z_{\max}}

\title{Comparing Linear Probes with Mahalanobis Cosine Similarity}

\author{
  Zhuofan Josh Ying\textsuperscript{1,4} \quad
  Peter Hase\textsuperscript{5,6} \quad
  Nikolaus Kriegeskorte\textsuperscript{1,2,3,4} \\[0.6em]
  Departments of \textsuperscript{1}Psychology, \textsuperscript{2}Neuroscience, \textsuperscript{3}Electrical Engineering \\
  \textsuperscript{4}Zuckerman Mind Brain Behavior Institute; Columbia University, New York, NY \\
  \textsuperscript{5}Stanford University, Stanford, CA \\
  \textsuperscript{6}Schmidt Sciences, New York, NY \\[0.4em]
  \texttt{zy2559@columbia.edu} \quad \texttt{phase@stanford.edu} \quad \texttt{nk2765@columbia.edu}
}

\begin{document}
\maketitle

\begin{abstract}
Linear probes are widely used in interpretability research and often compared by cosine similarity. The \emph{Mahalanobis cosine similarity} (MCS) between two directions, which reweights the inner product by test data covariance, is a natural task-aware refinement.
\citet{ying2026truthfulness} report that a probe's MCS to a reference probe trained on the out-of-distribution (OOD) data near-perfectly linearly predicts the probe's OOD AUROC ($R^2 {=} 0.98$).
Here, we extend this empirical finding across models, layers, and concept domains, and prove this general phenomenon in closed form: For balanced classes whose projections are Gaussian, OOD AUROC and MCS to the reference probe are linear because both are sigmoid-shaped functions of the probe's signal-to-noise ratio (SNR) on the test data. 
The theory also predicts when this linearity fails, which we verify empirically.
MCS offers a theoretically grounded and empirically effective alternative to Euclidean cosine similarity for comparing linear probes.
\end{abstract}

\section{Introduction}
\label{sec:intro}

Linear probes are powerful interpretability tools \citep{alain2016understanding, belinkov2022probing, marksgeometry}, but their transfer behavior is notoriously brittle: they often fail to generalize to closely related datasets, and break dramatically under subtle shifts such as negation or prompt-format changes \citep{hewitt2019designing, levinstein2024still, orgad2025llms}. 
Understanding which probes generalize, and why two probes that look similar on one distribution diverge on another, requires a principled way to compare them.



\citet{ying2026truthfulness} report that the \emph{Mahalanobis cosine similarity (MCS)} between a probe $\wid$ trained on an in-distribution (ID) task and another probe $\wood$ trained on an out-of-distribution (OOD) task, which reweights the inner product between the two probes by the OOD data covariance, is linearly related to the OOD AUROC of probe $\wid$ ($R^2 {=} 0.98$). 
However, this was shown only for one model on truthfulness datasets, and such near-perfect linearity demands a theoretical explanation.

Empirically, we replicate and substantially extend this regularity. The near-linear AUROC--MCS relationship holds across models (\texttt{Llama-70B}, \texttt{Llama-8B}, \texttt{Qwen-7B}), across layers (20--65 of \texttt{Llama-70B}), and across 24 datasets from three concept domains (truthfulness, gender classification, and general NLP benchmarks), with $R^2 {>} 0.93$ in every condition. Euclidean cosine similarity (ECS), by contrast, drops to as low as $R^2 {=} 0.06$ (\S\ref{sec:empirical}).
Intuitively, ECS treats all dimensions equally, but only the dimensions where the data varies affect probe performance. MCS reweights the inner product by OOD data covariance, thus comparing probes in the subspace that actually matters.



Theoretically, we prove why this linear relationship holds. 
Under balanced classes and per-class projection Gaussianity, both the ID probe's OOD AUROC and its MCS to the OOD probe are S-shapedd functions of the ID probe's signal-to-noise ratio (SNR) $s$ on the OOD task: AUROC is the Gaussian CDF $\Phi(s/\sqrt 2)$ in $s$, and MCS is a softsign function that converges to $s /\sqrt{(4+s^2)}$ with moderately large Fisher distance $\zmax$. 
Composing the two, where each S-shaped inverts the other's curvature, cancels out the S-shapeds, leaving a near-linear AUROC--MCS curve (\S\ref{sec:theory}). 
The theory also predicts when the linearity fails: small $\zmax$, class imbalance, non-Fisher reference directions like difference of means, or using pooled covariance instead of the total covariance. We verify each failure mode in simulations or on real activations (\S\ref{sec:failures}).

\begin{figure*}[t]
\centering
\includegraphics[width=0.98\linewidth]{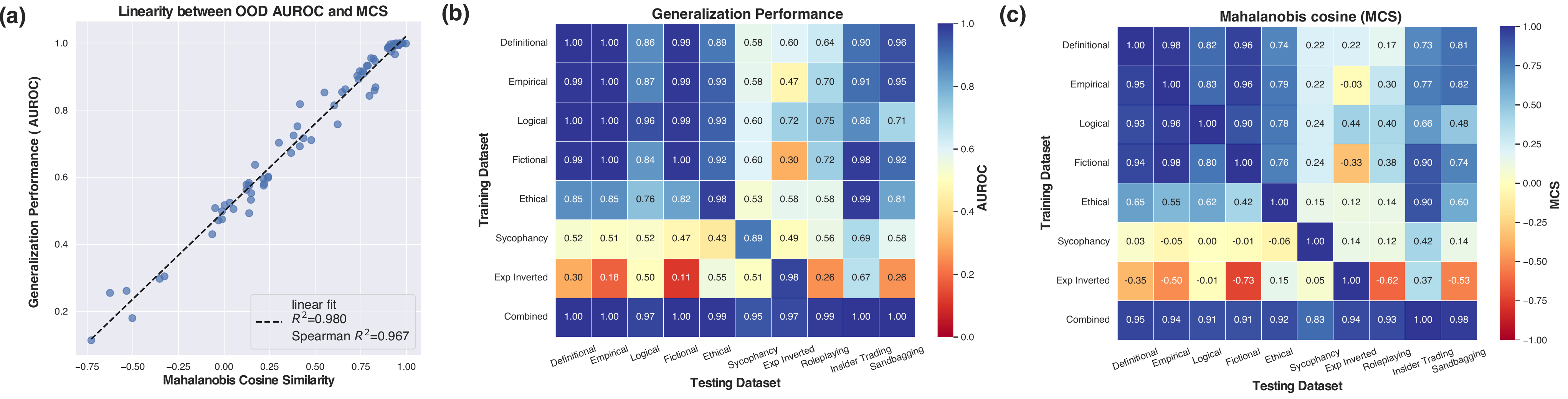}
\vspace{-10pt}
\caption{\textbf{Mahalanobis cosine similarity (MCS) linearly tracks generalization performance.} \emph{(a)} AUROC is a near-linear function of MCS across heterogeneous tasks. \emph{(b--c)} The generalization AUROC heatmap and the MCS heatmap share structure almost entry-for-entry. Reproduced from \citet{ying2026truthfulness}}
\label{fig:headline}
\vspace{-15pt}
\end{figure*}

Most elements here are textbook \citep{fisher1936use, green1966signal, chatfield2018introduction}, and MCS itself is not new \citep{bolme2003csu, iacovacci2020extraction}. The closest theoretical neighbor is Agreement-on-the-Line \citep{miller2021accuracy, baek2022agreement, baek2025theory}, which uses the same covariance-projected cosine to predict accuracy linearity between classifiers.  Composing these ingredients into a closed-form law relating MCS and OOD AUROC is novel. See App.~\ref{app:related} for extended related works.



Together, MCS offers a theoretically grounded task-aware alternative for comparing linear probes.

\begin{table}[t]
\centering\small
\begin{tabular}{lcc}
\toprule
Condition & $R^2$ (MCS) & $R^2$ (ECS) \\
\midrule
\texttt{Llama-70B}, L33, truth & 0.980 & 0.441 \\
\midrule
\multicolumn{3}{l}{\emph{Layers} (\texttt{Llama-70B}, truth)} \\
\quad layer 20 & 0.990 & 0.628 \\
\quad layer 50 & 0.963 & 0.422 \\
\quad layer 65 & 0.976 & 0.446 \\
\midrule
\multicolumn{3}{l}{\emph{Domains} (\texttt{Llama-70B}, L33)} \\
\quad Gender classification & 0.976 & 0.723 \\
\quad General NLP benchmarks & 0.936 & 0.057 \\
\midrule
\multicolumn{3}{l}{\emph{Models} (truth)} \\
\quad \texttt{Llama-3.1-8B}   & 0.935 & 0.530 \\
\quad \texttt{Qwen2.5-7B}     & 0.994 & 0.656 \\
\bottomrule
\end{tabular}
\caption{Linear-fit $R^2$ of held-out AUROC against Mahalanobis vs. standard Euclidean cosine similarity (MCS vs. ECS), across layers, domains, and model families and sizes. MCS dominates ECS in every condition.}
\label{tab:r2}
\vspace{-15pt}
\end{table}

\section{Empirical evidence}
\label{sec:empirical}

\paragraph{Design.}
We use \texttt{Llama-3.3-70B} at residual-stream layer 33, with mean-pooled activations over response tokens following \citet{ying2026truthfulness}. To test generality, we additionally evaluate at layers $\{20, 50, 65\}$ of \texttt{Llama-3.3-70B}, at middle layers of \texttt{Llama-3.1-8B} and \texttt{Qwen2.5-7B} \cite{qwen25, grattafiori2024llama}. We use datasets across three domains exhibiting rich generalization patterns: ten truthfulness datasets, six gender classification datasets, and eight general NLP classification datasets. See App.~\ref{app:exp-details} for details.

We train logistic-regression probes for each task. We denote the ID and OOD trained probe direction by $\wid$ and $\wood$. For directions $u,v$ we define
\vspace{-3.5mm}
\begin{equation}
    \MC_M(u,v) \coloneqq \frac{u^\top M v}{\sqrt{u^\top M u}\sqrt{v^\top M v}},
    \label{eq:mc-def}
\end{equation}
\vspace{-4.5mm}\\
and instantiate $M$ with the full sample OOD train data covariance $\Sigt$.\footnote{This is the opposite of the weighting in Mahalanobis distance (with $\Sigma^{-1}$), which measures distances between points. Since we instead care about distances between probe weights (dual space) rather than data points (primal space), our formulation transforms data and probes inversely and thus preserves projections onto the probe in the whitened space as desired.} 
We contrast against 
the pooled within-class covariance $\Sigp$ in \S\ref{sec:failures}. 
For each (ID, OOD) pair, we split the OOD data in half: on the \emph{train} half, we train the probes and compute $\Sigp$, $\Sigt$, $\delta$, $\wood$, and SNR; on the disjoint \emph{test} half, we compute the empirical AUROC of $\wid$. This avoids bias caused by overfitting.

\begin{figure*}[t]
\centering
\includegraphics[width=0.95\linewidth]{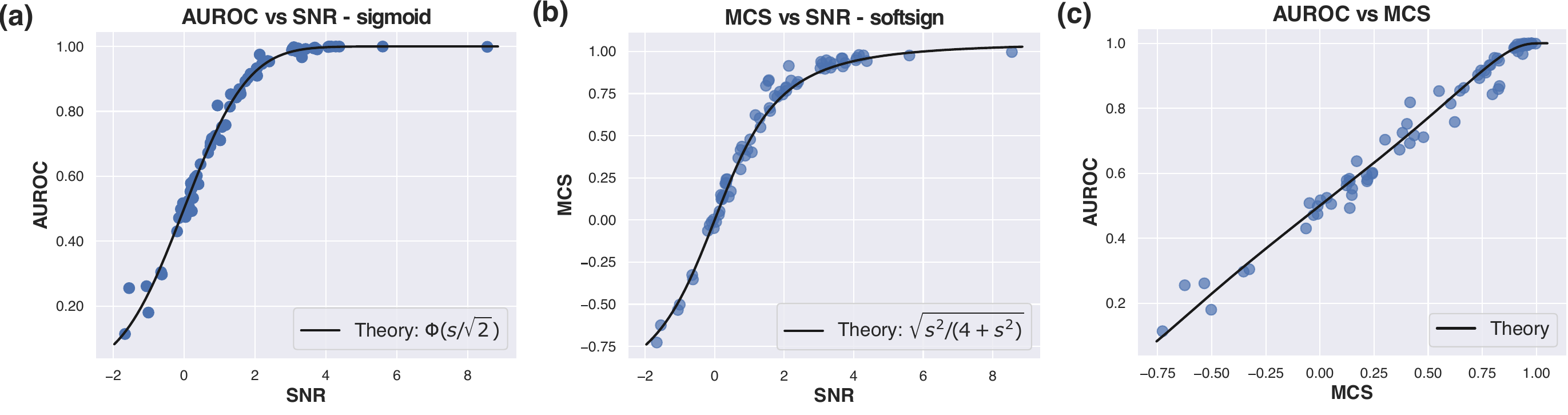}
\vspace{-8pt}
\caption{\textbf{Theory predicts empirical data without free parameters.} Across panels, empirical points largely lie on the theory prediction. \emph{(a)} AUROC--SNR shows Lemma~\ref{lem:auroc}. \emph{(b)} MCS--SNR shows Theorem~\ref{thm:mc}. \emph{(c)} 
Eliminating SNR, AUROC--MCS shows a near-straight line that bends only in the top-right corner, matching the empirical data.
}
\label{fig:proof}
\vspace{-15pt}
\end{figure*}

\paragraph{Results.}
We show that the OOD AUROC of $\wid$ is a near-linear function of $\MC_{\Sigt}(\wid, \wood)$, with $R^2 {\geq} 0.93$ for all conditions tested (Fig.~\ref{fig:headline}a, Tab.~\ref{tab:r2}). ECS is substantially worse: $R^2$ drops to 0.06 in the worst condition. Fig.~\ref{fig:headline}b--c shows that the structure of the generalization-AUROC heatmap (every (train, test) pair on the truthfulness benchmarks) is mirrored almost entry-for-entry by the MCS heatmap. 
See Fig.~\ref{fig:full_auroc},~\ref{fig:full_mc_vs_sc} and Tab.~\ref{tab:r2_lda} in App.~\ref{app:empirical} for the generalization performance, MCS and ECS plotted against AUROC, and robustness to $\Sigt$ estimators (full sample, Ledoit–Wolf, and per-coordinate diagonal) across all conditions.

\section{Theory}
\label{sec:theory}

\subsection{Setup}
\label{sec:theory-setup}

Let $X\in\R^d$ be a feature vector and $Y\in\{0,1\}$ a binary label. We analyse the OOD behaviour of a fixed direction $\wid$.

\begin{assumption}[Moments]\label{ass:moments}
Under the OOD distribution, $\Pr(Y{=}0){=}\Pr(Y{=}1){=}\tfrac12$, and the class-conditional distributions have means $\mu_0,\mu_1$ and covariances $\Sigma_0,\Sigma_1 \succ 0$. Define $\delta\coloneqq\mu_1{-}\mu_0$ and $\Sigp\coloneqq\tfrac12(\Sigma_0{+}\Sigma_1)$.
\end{assumption}

\begin{assumption}[Projection Gaussianity]\label{ass:gauss}
For each direction $w$ used in an AUROC statement below, the projection $w^\top X \mid Y{=}c$ is Gaussian for $c\in\{0,1\}$. This is much weaker than joint Gaussianity of $X$, and plausible even for non-Gaussian distributions in a high dimensional space by projection concentration (a Diaconis--Freedman-style CLT for one-dimensional projections).
\end{assumption}

Empirically, the per-class projections are largely normal (Fig.~\ref{fig:gaussian}), and simulations show the linearity survives clearly non-Gaussian projections (Fig.~\ref{fig:non-gaussian}). Since the empirical projections are Gaussian, we retain this assumption for simplicity.

For $w\in\R^d\setminus\{0\}$ we define
\vspace{-1mm}
\begin{equation}
\SNR(w) \coloneqq \frac{w^\top\delta}{\sqrt{w^\top\Sigp w}},
\zmax \coloneqq \sqrt{\delta^\top\Sigp^{-1}\delta},
\end{equation}
\vspace{-5mm}\\
In this section, $\wood$ denotes the Fisher discriminant direction $\wood {\coloneqq} \Sigp^{-1}\delta$, and we use $s {\coloneqq} \SNR(\wid)$. In experiments, we substitute the OOD-trained logistic-regression direction for the Fisher discriminant direction. 
Empirically, $\MC(\wid, \wood^{\mathrm{LR}}) {\approx} \MC(\wid, \wood^{\mathrm{LDA}})$ in all settings (see Fig.~\ref{fig:lr_vs_lda} in App.~\ref{app:lr-lda}), and the $R^2$ with AUROC are very close (see Tab.~\ref{tab:r2_lda} in App.~\ref{app:empirical}), so the substitution is largely harmless. 

\subsection{Background results}
\label{sec:theory-background}

The proof rests on three classic facts. We state them here and defer the textbook proofs to App.~\ref{app:classical-proofs}.

\begin{lemma}[Covariance decomposition]\label{lem:cov}
Under Assumption~\ref{ass:moments}, $\Sigt \coloneqq \Cov(X) = \Sigp + \tfrac14 \delta\delta^\top$.
\end{lemma}

\begin{lemma}[Binormal AUROC]\label{lem:auroc}
Under Assumptions~\ref{ass:moments}--\ref{ass:gauss}, $\AUROC(w) {=} \Phi\!\bigl(\SNR(w)/\sqrt 2\bigr)$,
where $\Phi$ is the standard Gaussian CDF, for any $w {\neq} 0$.
\end{lemma}


\begin{lemma}[Fisher's discriminant]\label{lem:fisher}
Under Assumption~\ref{ass:moments}, $\SNR^2(w)$ is maximised on $\R^d \setminus \{0\}$ by any nonzero scalar multiple of $\wood = \Sigp^{-1}\delta$, with maximum SNR $\zmax^2 = \delta^\top \Sigp^{-1} \delta$, where $\zmax$ is known as the Fisher distance.
\end{lemma}


\subsection{Closed form for the Mahalanobis cosine}
\label{sec:theory-mc}

The Mahalanobis cosine of an arbitrary direction $\wid$ with the Fisher direction $\wood$ admits a closed form in two scalars: the SNR of $\wid$ and the Fisher distance $\zmax$ of the task.

\begin{theorem}[Closed form for $\MC_{\Sigt}$]\label{thm:mc}
Under Assumption~\ref{ass:moments}, for any $\wid$ with $\wid^\top \Sigp \wid > 0$ and $\zmax > 0$,
\vspace{-3mm}
\begin{equation}
\MC_{\Sigt}(\wid,\wood)
= \frac{s}{\zmax}\sqrt{\frac{1+\tfrac14 \zmax^2}{1+\tfrac14 s^2}},
\label{eq:mc-tot}
\end{equation}
\vspace{-3mm}\\
where $s \coloneqq \SNR(\wid)$. Moreover, in the large-Fisher-distance limit,
\vspace{-3mm}
\begin{equation}
\lim_{\zmax \to \infty} \MC_{\Sigt}(s) \;=\; \frac{s}{\sqrt{4 + s^2}},
\label{eq:mc-softsign}
\end{equation}
\vspace{-4mm}\\
the softsign function: bounded in $(-1, 1)$, nearly linear near $s{=}0$, and saturating to $\pm 1$ as $|s| \to \infty$.
\end{theorem}

The proof is a direct computation of the three quadratic forms in~\eqref{eq:mc-def} using Lemma~\ref{lem:cov} (App.~\ref{app:thm-proof}). Although the derivation uses no objects beyond those already in the LDA literature, the resulting identity has not been written down, likely because the question it answers — how MCS of an arbitrary direction with the Fisher direction depends on SNR — was not previously connected to probe transfer.  


The algebra is basic; what matters is the \emph{shape}. On $(-\zmax, \zmax)$, $\MC_{\Sigt}(s)$ is odd, strictly increasing, and satisfies $\MC_{\Sigt}(\pm\zmax) {=} \pm 1$. As $\zmax {\to} \infty$, $\MC_{\Sigt}(s)$ is exactly a softsign shape (Eq.~\ref{eq:mc-softsign}; Fig.~\ref{fig:proof}(b)). 

\subsection{Why AUROC is linear in MCS, with task-independent slope}
\label{sec:theory-slope}
 
Lemma~\ref{lem:auroc} and Theorem~\ref{thm:mc} together imply that for a fixed OOD task (fixed $\zmax$), both $\AUROC$ and $\MC_{\Sigt}$ are monotone functions of the signal-to-noise ratio $s$, tracing a parametric curve indexed only by $\zmax$. 
Two facts make this curve a near-linear law with a universal slope.
 
\paragraph{(i) Fixed $\zmax$: saturations cancel.}
Differentiating the parametric curve at any $s$ gives:
\vspace{-3mm}
\begin{equation}
\frac{d\,\AUROC}{d\,\MC_{\Sigt}}
\;=\; \frac{\phi(s/\sqrt 2)\,(1+s^2/4)^{3/2}}{\sqrt{2/\zmax^2+1/2}},
\label{eq:slope-full}
\end{equation}
\vspace{-5mm}\\
where $\phi$ is the standard Gaussian density (App.~\ref{app:slope}). The two $s$-dependent factors are \emph{opposed} in $|s|$: as $|s|$ grows, $\phi(s/\sqrt 2)$ \emph{shrinks} (the AUROC sigmoid saturating) while $(1{+}s^2/4)^{3/2}$ \emph{grows} (the MCS softsign saturating). Their product is therefore much flatter than either factor alone, so the local slope stays close to its central value over the bulk of $\MC_{\Sigt}\in(-1,1)$, as shown in Fig.~\ref{fig:proof}c.
 
The cancellation is not exact: as $|\MC_{\Sigt}|\to 1$, the Gaussian factor decays faster than the polynomial grows, and the local slope drops toward $0$ (both data and theory at top-right of Fig.~\ref{fig:proof}c flatten).
 
\paragraph{(ii) Per-task slope is universal.}
The slope in Eq.~\ref{eq:slope-full} factors as $h(s)\cdot g(\zmax)$, where $h(s)$ captures $s$-dependence and $g(\zmax) {=} 1/\sqrt{2/\zmax^2 + 1/2}$ captures task dependence. 
At $\zmax {>} 20$, which holds in all empirical conditions (Tab.~\ref{tab:zmax}), $g(\zmax)$ lies within 0.5\% of its limit. Therefore, the per-task AUROC--MCS curve aligns with the $\zmax{\to}\infty$ limit curve to within 0.5\%. 
At $s{=}0$, this limit curve has central slope $h(0)\cdot g(\infty){=}1/\sqrt{\pi}$ (App.~\ref{app:slope}).
Combined with (i), which establishes that the limit curve is approximately linear, with empirical slopes lying slightly below $1/\sqrt{\pi}$ as expected (Fig.~\ref{fig:slope}; App.~\ref{app:slope}), since saturation-tail sampling drags the global slope down from its central value.

 

\section{When does the linearity break down}
\label{sec:failures}

\begin{figure}[t]
\centering
\includegraphics[width=0.98\linewidth]{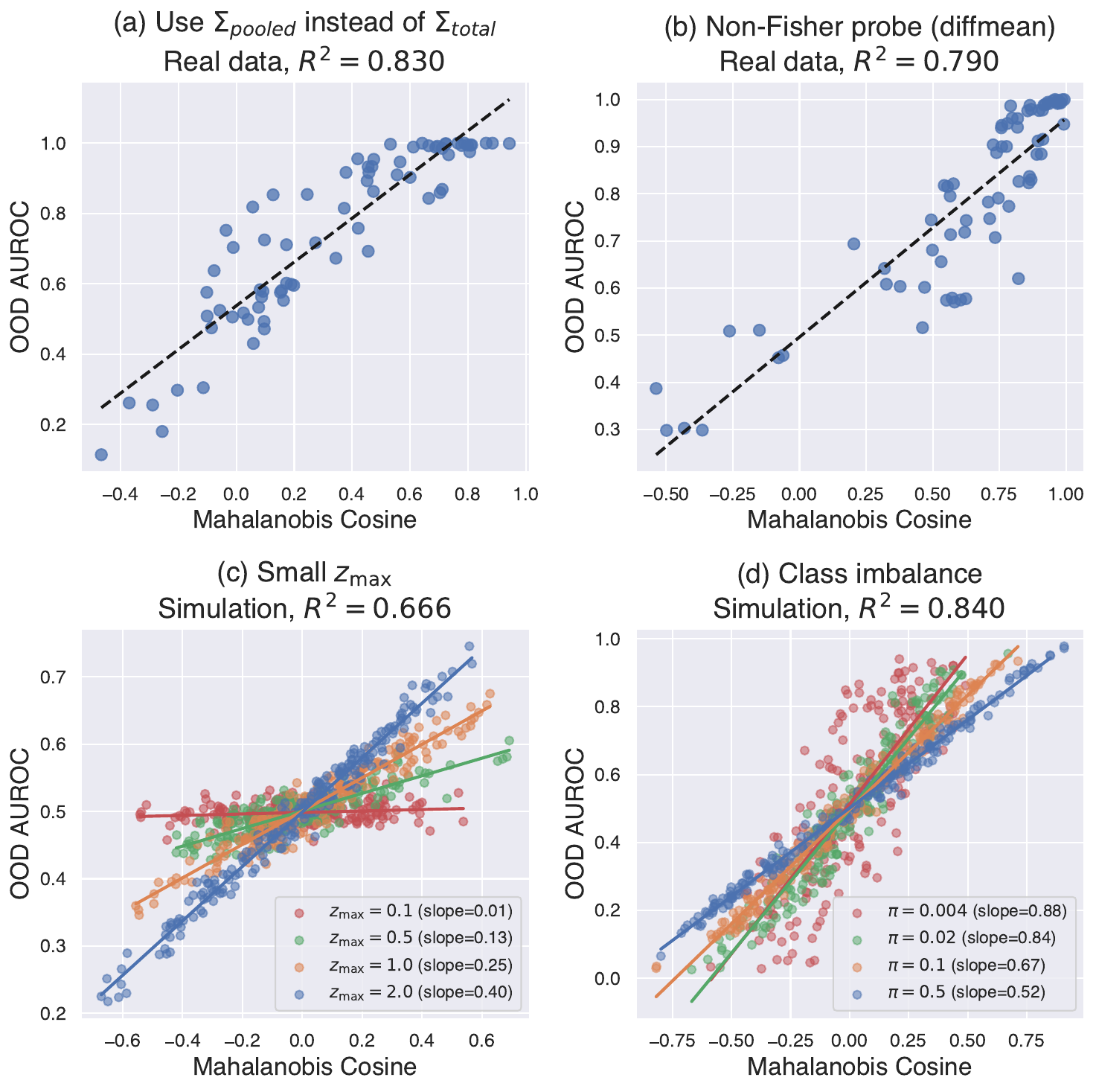}
\vspace{-10pt}
\caption{\textbf{Failure modes.} Each panel illustrates a violation of an assumption in \S\ref{sec:theory}, and the linearity breaks.}
\label{fig:failures}
\vspace{-16pt}
\end{figure}

The AUROC--MCS linearity holds when: MCS is computed against $\Sigt$, not $\Sigp$; the Fisher distance $\zmax$ is large; class proportions are roughly balanced; the OOD probes are close to the optimal Fisher direction. Each ingredient can fail in practice. Fig.~\ref{fig:failures} shows what each violation does to the linearity; the four cases are quoted at the same axes throughout, and reported $R^2$ are linear fits of OOD AUROC against $\MC$. 

\textbf{(a) Wrong covariance.}
Using $\Sigp$ instead of $\Sigt$ turns $\MC$ into $s/\zmax$ exactly, so AUROC becomes a sigmoid in MCS rather than a line (see App.~\ref{app:failures} for derivation). On the main experiment data, the linear fit $R^2$ drops from 0.98 to 0.83. 

\textbf{(b) Non-Fisher probe.} The theory does not hold if OOD probes deviate too much from the optimal Fisher directions, for example, for the commonly used difference of means probe \cite{marksgeometry}. On the same empirical data, the diffmean probe gives a markedly lower $R^2$ of 0.79. This delimits the law: it predicts generalization for Fisher-style probes (LR, LDA, shrinkage variants), not for diffmean-style probes. 

\textbf{(c) Small Fisher distance.}
For small $\zmax$, the slope of the MCS formula does not saturate, so each task is in its own near-linear regime with its own slope. Synthetic Gaussian data at $\zmax \in \{0.1, 0.5, 1, 2\}$ exhibits a clear fan of per-group slopes spanning a large range. The LLM data is unaffected because $\zmax {>} 20$ on every tasks (see Tab.~\ref{tab:zmax}; App.~\ref{app:empirical}). The linear fit is tight for each group, but taken together, the $R^2$ drops to 0.666. 

\textbf{(d) Class imbalance.}
The balanced-class factor $\tfrac14$ in the MCS formula generalises to $\pi(1-\pi)$, so the AUROC--MCS slope steepens as $\pi$ moves away from $\tfrac12$ (App.~\ref{app:failures}). Synthetic data at $\pi \in \{0.5, 0.1, 0.02, 0.004\}$ shows monotone steepening. The $R^2$ drops to 0.84 in this case.

\section{Discussion}
\label{sec:discussion}

Our results suggest that Mahalanobis cosine similarity is a theoretically sound alternative to standard Euclidean cosine similarity (consistent with recent arguments for non-Euclidean inner products \citep{park2024linear}). 
This also points to a broader research direction: many interpretability methods that currently rely on cosine similarity may benefit from the Mahalanobis alternative (steering-vector comparison, SAE feature alignment, concept-direction clustering, data filtering, data attribution, etc.).
Future work can use unsupervised methods like CCS \citep{burns2023discovering} for the OOD direction, enabling fully label-free prediction of probe generalization.

\clearpage

\section*{Limitations}
Our results have several limitations. First, computing $\mathrm{MC}_{\Sigma_{\mathrm{tot}}}$ requires estimating $w_{\mathrm{ood}}$ and $\Sigma_{\mathrm{tot}}$ from labeled OOD data; the method is training-free for the candidate probe $w_{\mathrm{id}}$ but not label-free for the target distribution. Second, the theory is calibrated to Fisher-style probes (LR, LDA); for difference-of-means probes the linearity degrades markedly (Fig.~\ref{fig:failures}b), and a closed form for non-Fisher references remains open. Third, our evaluation is confined to binary classification on residual-stream features of autoregressive LLMs; multiclass probes, attention/MLP features, and non-LLM architectures are untested. Fourth, the law is fundamentally approximate: the per-task total slope is close to $1/\sqrt{\pi}$ and the local slope goes to 0 at the extreme. Finally, the theory assumes per-class projection Gaussianity; while this holds empirically for most directions (Fig.~\ref{fig:gaussian}) and the linearity survives clear violations in simulation (Fig.~\ref{fig:non-gaussian}), formal guarantees do not extend to heavy-tailed projections.
\paragraph{Risks.} We do not foresee risks specific to this work; it is a theoretical analysis of an existing similarity measure.
\bibliography{custom}

@inproceedings{marksgeometry,
  title={The Geometry of Truth: Emergent Linear Structure in Large Language Model Representations of True/False Datasets},
  author={Marks, Samuel and Tegmark, Max},
  booktitle={First Conference on Language Modeling},
year={2023}
}

@inproceedings{goldowsky2025detecting,
  title={Detecting strategic deception with linear probes},
  author={Goldowsky-Dill, Nicholas and Chughtai, Bilal and Heimersheim, Stefan and Hobbhahn, Marius},
  booktitle={Forty-second International Conference on Machine Learning},
  year={2025}
}

@inproceedings{burns2023discovering,
  title={Discovering Latent Knowledge in Language Models Without Supervision},
  author={Burns, Collin and Ye, Haotian and Klein, Dan and Steinhardt, Jacob},
  booktitle={The Eleventh International Conference on Learning Representations},
year={2023}
}

@inproceedings{scheurer2024insider_trading,
  title={Large Language Models can Strategically Deceive their Users when Put Under Pressure},
  author={Scheurer, J{\'e}r{\'e}my and Balesni, Mikita and Hobbhahn, Marius},
  booktitle={ICLR 2024 Workshop on Large Language Model (LLM) Agents},
    year={2023}
}

@article{benton2024sabotage,
  title={Sabotage evaluations for frontier models},
  author={Benton, Joe and Wagner, Misha and Christiansen, Eric and Anil, Cem and Perez, Ethan and Srivastav, Jai and Durmus, Esin and Ganguli, Deep and Kravec, Shauna and Shlegeris, Buck and others},
  journal={arXiv preprint arXiv:2410.21514},
  year={2024}
}

@article{grattafiori2024llama,
  title={The llama 3 herd of models},
  author={Grattafiori, Aaron and Dubey, Abhimanyu and Jauhri, Abhinav and Pandey, Abhinav and Kadian, Abhishek and Al-Dahle, Ahmad and Letman, Aiesha and Mathur, Akhil and Schelten, Alan and Vaughan, Alex and others},
  journal={arXiv preprint arXiv:2407.21783},
  year={2024}
}

@misc{qwen25,
      title={Qwen2.5 Technical Report}, 
      author={Qwen and An Yang and Baosong Yang and Beichen Zhang and Binyuan Hui and Bo Zheng and Bowen Yu and Chengyuan Li and Dayiheng Liu and Fei Huang and Haoran Wei and Huan Lin and Jian Yang and Jianhong Tu and Jianwei Zhang and Jianxin Yang and Jiaxi Yang and Jingren Zhou and Junyang Lin and Kai Dang and Keming Lu and Keqin Bao and Kexin Yang and Le Yu and Mei Li and Mingfeng Xue and Pei Zhang and Qin Zhu and Rui Men and Runji Lin and Tianhao Li and Tianyi Tang and Tingyu Xia and Xingzhang Ren and Xuancheng Ren and Yang Fan and Yang Su and Yichang Zhang and Yu Wan and Yuqiong Liu and Zeyu Cui and Zhenru Zhang and Zihan Qiu},
      year={2025},
      eprint={2412.15115},
      archivePrefix={arXiv},
      primaryClass={cs.CL},
      url={https://arxiv.org/abs/2412.15115}, 
}

@inproceedings{orgad2025llms,
  title={LLMs Know More Than They Show: On the Intrinsic Representation of LLM Hallucinations},
  author={Orgad, Hadas and Toker, Michael and Gekhman, Zorik and Reichart, Roi and Szpektor, Idan and Kotek, Hadas and Belinkov, Yonatan},
  booktitle={ICLR},
  year={2025}
}

@article{levinstein2024still,
  title={Still no lie detector for language models: probing empirical and conceptual roadblocks},
  author={Levinstein, Benjamin A and Herrmann, Daniel A},
  journal={Philosophical Studies},
  year={2024},
  publisher={Springer Netherlands}
}

@inproceedings{wolf2020transformers,
  title={Transformers: State-of-the-art natural language processing},
  author={Wolf, Thomas and Debut, Lysandre and Sanh, Victor and Chaumond, Julien and Delangue, Clement and Moi, Anthony and Cistac, Pierric and Rault, Tim and Louf, Remi and Funtowicz, Morgan and others},
  booktitle={Proceedings of the 2020 conference on empirical methods in natural language processing: system demonstrations},
  pages={38--45},
  year={2020}
}

@inproceedings{nnsight,
  title={NNsight and NDIF: Democratizing Access to Open-Weight Foundation Model Internals},
  author={Fiotto-Kaufman, Jaden Fried and Loftus, Alexander Russell and Todd, Eric and Brinkmann, Jannik and Pal, Koyena and Troitskii, Dmitrii and Ripa, Michael and Belfki, Adam and Rager, Can and Juang, Caden and others},
  booktitle={The Thirteenth International Conference on Learning Representations},
year={2024}
}

@article{belinkov2022probing,
  title={Probing classifiers: Promises, shortcomings, and advances},
  author={Belinkov, Yonatan},
  journal={Computational Linguistics},
  volume={48},
  number={1},
  pages={207--219},
  year={2022}
}

@inproceedings{you2021logme,
  title={Logme: Practical assessment of pre-trained models for transfer learning},
  author={You, Kaichao and Liu, Yong and Wang, Jianmin and Long, Mingsheng},
  booktitle={International conference on machine learning},
  pages={12133--12143},
  year={2021},
  organization={PMLR}
}

@article{pedregosa2011scikit,
  title={Scikit-learn: Machine learning in Python},
  author={Pedregosa, Fabian and Varoquaux, Ga{\"e}l and Gramfort, Alexandre and Michel, Vincent and Thirion, Bertrand and Grisel, Olivier and Blondel, Mathieu and Prettenhofer, Peter and Weiss, Ron and Dubourg, Vincent and others},
  journal={the Journal of machine Learning research},
  volume={12},
  pages={2825--2830},
  year={2011},
  publisher={JMLR. org}
}

@article{ying2026truthfulness,
  title={The Truthfulness Spectrum Hypothesis},
  author={Ying, Zhuofan Josh and Ravfogel, Shauli and Kriegeskorte, Nikolaus and Hase, Peter},
  journal={arXiv preprint arXiv:2602.20273},
  year={2026}
}

@article{alain2016understanding,
  title={Understanding intermediate layers using linear classifier probes},
  author={Alain, Guillaume and Bengio, Yoshua},
  journal={arXiv preprint arXiv:1610.01644},
  year={2016}
}

@inproceedings{bolme2003csu,
  title={The CSU face identification evaluation system: its purpose, features, and structure},
  author={Bolme, David S and Ross Beveridge, J and Teixeira, Marcio and Draper, Bruce A},
  booktitle={International Conference on Computer Vision Systems},
  pages={304--313},
  year={2003},
  organization={Springer}
}

@article{fisher1936use,
  title={The use of multiple measurements in taxonomic problems},
  author={Fisher, Ronald A},
  journal={Annals of eugenics},
  volume={7},
  number={2},
  pages={179--188},
  year={1936},
  publisher={Wiley Online Library}
}

@book{green1966signal,
  title={Signal detection theory and psychophysics},
  author={Green, David Marvin and Swets, John A and others},
  volume={1},
  year={1966},
  publisher={Wiley New York}
}

@article{hanley1982meaning,
  title={The meaning and use of the area under a receiver operating characteristic (ROC) curve.},
  author={Hanley, James A and McNeil, Barbara J},
  journal={Radiology},
  volume={143},
  number={1},
  pages={29--36},
  year={1982}
}

@book{chatfield2018introduction,
  title={Introduction to multivariate analysis},
  author={Chatfield, Chris},
  year={2018},
  publisher={Routledge}
}

@misc{hastie2009elements,
  title={The elements of statistical learning: data mining, inference, and prediction},
  author={Hastie, Trevor},
  year={2009},
  publisher={springer}
}

@inproceedings{hewitt2019designing,
  title={Designing and interpreting probes with control tasks},
  author={Hewitt, John and Liang, Percy},
  booktitle={Proceedings of the 2019 conference on empirical methods in natural language processing and the 9th international joint conference on natural language processing (emnlp-ijcnlp)},
  pages={2733--2743},
  year={2019}
}

@article{iacovacci2020extraction,
  title={Extraction and integration of genetic networks from short-profile omic data sets},
  author={Iacovacci, Jacopo and Peluso, Alina and Ebbels, Timothy and Ralser, Markus and Glen, Robert C},
  journal={Metabolites},
  volume={10},
  number={11},
  pages={435},
  year={2020},
  publisher={MDPI}
}

@inproceedings{pandy2022transferability,
  title={Transferability estimation using bhattacharyya class separability},
  author={P{\'a}ndy, Michal and Agostinelli, Andrea and Uijlings, Jasper and Ferrari, Vittorio and Mensink, Thomas},
  booktitle={Proceedings of the IEEE/CVF Conference on Computer Vision and Pattern Recognition},
  pages={9172--9182},
  year={2022}
}

@inproceedings{maas2011learning,
  title={Learning word vectors for sentiment analysis},
  author={Maas, Andrew and Daly, Raymond E and Pham, Peter T and Huang, Dan and Ng, Andrew Y and Potts, Christopher},
  booktitle={Proceedings of the 49th annual meeting of the association for computational linguistics: Human language technologies},
  pages={142--150},
  year={2011}
}

@inproceedings{mcauley2013hidden,
  title={Hidden factors and hidden topics: understanding rating dimensions with review text},
  author={McAuley, Julian and Leskovec, Jure},
  booktitle={Proceedings of the 7th ACM conference on Recommender systems},
  pages={165--172},
  year={2013}
}

@article{zhang2015character,
  title={Character-level convolutional networks for text classification},
  author={Zhang, Xiang and Zhao, Junbo and LeCun, Yann},
  journal={Advances in neural information processing systems},
  volume={28},
  year={2015}
}

@article{lehmann2015dbpedia,
  title={Dbpedia--a large-scale, multilingual knowledge base extracted from wikipedia},
  author={Lehmann, Jens and Isele, Robert and Jakob, Max and Jentzsch, Anja and Kontokostas, Dimitris and Mendes, Pablo N and Hellmann, Sebastian and Morsey, Mohamed and Van Kleef, Patrick and Auer, S{\"o}ren and others},
  journal={Semantic web},
  volume={6},
  number={2},
  pages={167--195},
  year={2015},
  publisher={SAGE Publications Sage UK: London, England}
}

@inproceedings{wang2018glue,
  title={GLUE: A multi-task benchmark and analysis platform for natural language understanding},
  author={Wang, Alex and Singh, Amanpreet and Michael, Julian and Hill, Felix and Levy, Omer and Bowman, Samuel},
  booktitle={Proceedings of the 2018 EMNLP workshop BlackboxNLP: Analyzing and interpreting neural networks for NLP},
  pages={353--355},
  year={2018}
}

@inproceedings{clark2019boolq,
  title={Boolq: Exploring the surprising difficulty of natural yes/no questions},
  author={Clark, Christopher and Lee, Kenton and Chang, Ming-Wei and Kwiatkowski, Tom and Collins, Michael and Toutanova, Kristina},
  booktitle={Proceedings of the 2019 conference of the north American chapter of the association for computational linguistics: Human language technologies, volume 1 (long and short papers)},
  pages={2924--2936},
  year={2019}
}

@inproceedings{bisk2020piqa,
  title={Piqa: Reasoning about physical commonsense in natural language},
  author={Bisk, Yonatan and Zellers, Rowan and Gao, Jianfeng and Choi, Yejin and others},
  booktitle={Proceedings of the AAAI conference on artificial intelligence},
  volume={34},
  pages={7432--7439},
  year={2020}
}

@inproceedings{park2024linear,
  title={The linear representation hypothesis and the geometry of large language models},
  author={Park, Kiho and Choe, Yo Joong and Veitch, Victor},
  booktitle={Proceedings of the 41st International Conference on Machine Learning},
  pages={39643--39666},
  year={2024}
}

@inproceedings{martinc2017pan,
  title={PAN 2017: Author Profiling-Gender and Language Variety Prediction.},
  author={Martinc, Matej and Skrjanec, Iza and Zupan, Katja and Pollak, Senja},
  booktitle={CLEF (working notes)},
  year={2017}
}

@article{eight2016twitter,
  title={Twitter user gender classification},
  author={Eight, Figure},
  journal={Kaggle Dataset},
  year={2016},
  url={https://www.kaggle.com/datasets/crowdflower/twitter-user-gender-classification}
}

@inproceedings{asai2018happydb,
  title={Happydb: A corpus of 100,000 crowdsourced happy moments},
  author={Asai, Akari and Evensen, Sara and Golshan, Behzad and Halevy, Alon and Li, Vivian and Lopatenko, Andrei and Stepanov, Daniela and Suhara, Yoshi and Tan, Wang-Chiew and Xu, Yinzhan},
  booktitle={Proceedings of the Eleventh International Conference on Language Resources and Evaluation (LREC 2018)},
  year={2018}
}

@inproceedings{de2019bias,
  title={Bias in bios: A case study of semantic representation bias in a high-stakes setting},
  author={De-Arteaga, Maria and Romanov, Alexey and Wallach, Hanna and Chayes, Jennifer and Borgs, Christian and Chouldechova, Alexandra and Geyik, Sahin and Kenthapadi, Krishnaram and Kalai, Adam Tauman},
  booktitle={proceedings of the Conference on Fairness, Accountability, and Transparency},
  pages={120--128},
  year={2019}
}

@inproceedings{zhao2018gender,
  title={Gender bias in coreference resolution: Evaluation and debiasing methods},
  author={Zhao, Jieyu and Wang, Tianlu and Yatskar, Mark and Ordonez, Vicente and Chang, Kai-Wei},
  booktitle={Proceedings of the 2018 Conference of the North American Chapter of the Association for Computational Linguistics: Human Language Technologies, Volume 2 (Short Papers)},
  pages={15--20},
  year={2018}
}

@article{webster2018mind,
  title={Mind the GAP: A balanced corpus of gendered ambiguous pronouns},
  author={Webster, Kellie and Recasens, Marta and Axelrod, Vera and Baldridge, Jason},
  journal={Transactions of the Association for Computational Linguistics},
  volume={6},
  pages={605--617},
  year={2018},
  publisher={MIT Press One Rogers Street, Cambridge, MA 02142-1209, USA journals-info~…}
}

@article{efron1975efficiency,
  title={The efficiency of logistic regression compared to normal discriminant analysis},
  author={Efron, Bradley},
  journal={Journal of the American Statistical Association},
  volume={70},
  number={352},
  pages={892--898},
  year={1975},
  publisher={Taylor \& Francis}
}

@article{cristianini2001kernel,
  title={On kernel-target alignment},
  author={Cristianini, Nello and Shawe-Taylor, John and Elisseeff, Andre and Kandola, Jaz},
  journal={Advances in neural information processing systems},
  volume={14},
  year={2001}
}

@inproceedings{garrido2023rankme,
  title={Rankme: Assessing the downstream performance of pretrained self-supervised representations by their rank},
  author={Garrido, Quentin and Balestriero, Randall and Najman, Laurent and Lecun, Yann},
  booktitle={International conference on machine learning},
  pages={10929--10974},
  year={2023},
  organization={PMLR}
}

@article{papyan2020prevalence,
  title={Prevalence of neural collapse during the terminal phase of deep learning training},
  author={Papyan, Vardan and Han, XY and Donoho, David L},
  journal={Proceedings of the National Academy of Sciences},
  volume={117},
  number={40},
  pages={24652--24663},
  year={2020},
  publisher={National Academy of Sciences}
}

@inproceedings{galanti2021role,
  title={On the Role of Neural Collapse in Transfer Learning},
  author={Galanti, Tomer and Gy{\"o}rgy, Andr{\'a}s and Hutter, Marcus},
  booktitle={International Conference on Learning Representations},
  year={2021}
}

@article{lee2018simple,
  title={A simple unified framework for detecting out-of-distribution samples and adversarial attacks},
  author={Lee, Kimin and Lee, Kibok and Lee, Honglak and Shin, Jinwoo},
  journal={Advances in neural information processing systems},
  volume={31},
  year={2018}
}

@article{ren2021simple,
  title={A simple fix to mahalanobis distance for improving near-ood detection},
  author={Ren, Jie and Fort, Stanislav and Liu, Jeremiah and Roy, Abhijit Guha and Padhy, Shreyas and Lakshminarayanan, Balaji},
  journal={arXiv preprint arXiv:2106.09022},
  year={2021}
}

@inproceedings{miller2021accuracy,
  title={Accuracy on the line: on the strong correlation between out-of-distribution and in-distribution generalization},
  author={Miller, John P and Taori, Rohan and Raghunathan, Aditi and Sagawa, Shiori and Koh, Pang Wei and Shankar, Vaishaal and Liang, Percy and Carmon, Yair and Schmidt, Ludwig},
  booktitle={International conference on machine learning},
  pages={7721--7735},
  year={2021},
  organization={PMLR}
}

@article{baek2022agreement,
  title={Agreement-on-the-line: Predicting the performance of neural networks under distribution shift},
  author={Baek, Christina and Jiang, Yiding and Raghunathan, Aditi and Kolter, J Zico},
  journal={Advances in Neural Information Processing Systems},
  volume={35},
  pages={19274--19289},
  year={2022}
}

@inproceedings{baek2025theory,
  title={Theory of Agreement-on-the-Line in Linear Models and Gaussian Data},
  author={Baek, Christina and Raghunathan, Aditi and Kolter, J Zico},
  booktitle={The 28th International Conference on Artificial Intelligence and Statistics},
  year={2025}
}

@article{bamber1975area,
  title={The area above the ordinal dominance graph and the area below the receiver operating characteristic graph},
  author={Bamber, Donald},
  journal={Journal of mathematical psychology},
  volume={12},
  number={4},
  pages={387--415},
  year={1975},
  publisher={Elsevier}
}

@article{wilks1932certain,
  title={Certain generalizations in the analysis of variance},
  author={Wilks, Samuel S},
  journal={Biometrika},
  volume={24},
  number={3/4},
  pages={471--494},
  year={1932},
  publisher={JSTOR}
}

\clearpage
\appendix

\section{Extended related works}
\label{app:related}

\paragraph{Linear probes and their generalization.}
Linear probes are a standard interpretability tool \citep{alain2016understanding, belinkov2022probing}, used to read off concepts such as truthfulness \citep{marksgeometry}, but often degrade under distribution shift \citep{hewitt2019designing, levinstein2024still, orgad2025llms}. Prior work predicts linear-classifier generalization from geometric properties of the representation---kernel-target alignment \citep{cristianini2001kernel}, spectral rank \citep{garrido2023rankme}, and neural-collapse statistics \citep{papyan2020prevalence, galanti2021role}---but these score whole representations rather than individual probe directions. The near-linear AUROC--MCS relationship we explain was reported empirically by \citet{ying2026truthfulness} on truthfulness datasets; we prove it, characterize when it fails, and verify it across models, layers, and domains.

\paragraph{Classical statistical machinery.}
Our theory composes three textbook results: Fisher's linear discriminant \citep{fisher1936use}, the binormal AUROC formula \citep{bamber1975area, green1966signal, hanley1982meaning}, and the within--between covariance decomposition underlying MANOVA \citep{wilks1932certain, chatfield2018introduction}. None is novel in isolation; the contribution lies in composing them into a closed-form quantity computable from a candidate probe direction together with two target-task moments.

\paragraph{Agreement-on-the-Line.} The closest theoretical neighbor is the Agreement-on-the-Line framework \citep{miller2021accuracy,baek2022agreement,baek2025theory}, which proves a linear ID-to-OOD relationship for probit-scaled accuracy and agreement rates between linear classifiers under Gaussian data. \citet{baek2025theory} parameterize this relationship by a similarity that is algebraically the Mahalanobis cosine of Eq.~(1). We differ in (i) targeting threshold-free AUROC, which admits the closed form of Theorem~1; (ii) predicting OOD performance from a single geometric quantity against the OOD Fisher direction, with no ID baseline; and (iii) identifying the universal slope around $1/\sqrt{\pi}$ from the AUROC--MCS saturation cancellation, which has no analog in the AOL line.

\paragraph{Mahalanobis cosine as a similarity for direction comparison.}
The most direct precursor of our setup is \citet{bolme2003csu}, who introduced "Mahalanobis cosine" as a similarity measure in PCA-whitened face-recognition pipelines. \citet{iacovacci2020extraction} reused the same expression for inferring omics networks. Both works treat MCS as a similarity score to be evaluated empirically against task performance; neither expresses MCS in closed form as a function of probe SNR and task Fisher distance, and neither links it to AUROC. Our contribution in \S\ref{sec:theory} is to make precise why MCS works as a direction-comparison similarity for classification — in a closed form (Theorem 1) whose composition with the binormal AUROC is task-calibrated and predicts held-out performance.

\paragraph{Non-Euclidean geometry of LLM representations.}
\citet{park2024linear} also argue against Euclidean cosine on LLM representation space and propose a "causal inner product" derived from counterfactual considerations on the unembedding space. Their specific instantiation whitens by the covariance of unembedding token vectors, a single global matrix per model. Our $\Sigma_{tot}$ is a per-task activation covariance computed on the OOD train data, and operates in the residual-stream embedding space rather than the unembedding space. 

\paragraph{Mahalanobis distance in OOD detection and transferability.}
Mahalanobis distances of test points are widely used as out-of-distribution scores \citep{lee2018simple, ren2021simple}, but they score raw inputs rather than candidate probe directions. A separate transferability-metrics literature (LEEP, LogME, H-score, GBC, NCTI, Task2Vec; e.g.\ \citealp{you2021logme, pandy2022transferability}) predicts downstream linear-probe performance from frozen features, but operates on whole representations and typically requires target-side labels. To our knowledge, no prior method predicts the held-out AUROC of a specific probe direction in closed form from two target-task moments.

\section{Experimental details}
\label{app:exp-details}

\subsection{Computational resources.} 
All experiments are done on local L40S and A40. Experiments require 4 GPUs to run for about 10 hours. We use \texttt{Huggingface} and \texttt{NNsight} to extract activations \cite{wolf2020transformers, nnsight}. The logistic regression probe is implemented with \citet{pedregosa2011scikit}. 

\subsection{Model and probe details}

\paragraph{Models.} We use \texttt{Llama-3.3-70B}, \texttt{Llama-3.1-8B}, and \texttt{Qwen-2.5-7B} \cite{grattafiori2024llama, qwen25}. We use layer 15 and layer 19 for \texttt{Llama-3.1-8B} and \texttt{Qwen-2.5-7B} following \citet{ying2026truthfulness}.

\subsection{Dataset details}

\paragraph{Truthfulness datasets.} We use ten total truthfulness datasets: five fundamental truth types from the FLEED dataset (definitional, empirical, logical, fictional, and ethical truth) ($\approx$8k samples), sycophantic lying ($\approx$4k samples), expectation inverted lying ($\approx$600 sample), and three on-policy deception datasets (roleplay (371 samples), insider trading (1.5k samples), and sandbagging (1k samples)) \cite{scheurer2024insider_trading, benton2024sabotage, goldowsky2025detecting, ying2026truthfulness}. For \texttt{Qwen-2.5-7B}, we don't have the three on-policy deception datasets. Therefore, it is evaluated only on 7 datasets. 

\paragraph{Gender classification datasets.} We use six gender classification datasets: PAN17 \cite{martinc2017pan}, CrowdFlower Twitter gender classification \cite{eight2016twitter}, HappyDB \cite{asai2018happydb}, BiosBias \cite{de2019bias}, WinoBias \cite{zhao2018gender}, and GAP \cite{webster2018mind}. For datasets larger than 4000 samples, we randomly subsample to 4000 samples. Therefore, GAP has 2,000 samples, PAN17 randomly combines $N$ tweets from the same user and has 2,400 samples, WinoBias has 1,584 samples, and the rest has 4,000 samples each.

\paragraph{General NLP benchmarks.} We use eight classic NLP QA datasets: sentiment classification (IMDB \cite{maas2011learning} and Amazon \cite{mcauley2013hidden}), topic classification (AG-News \cite{zhang2015character} and DBpedia-14 \cite{lehmann2015dbpedia}), NLI (RTE \cite{wang2018glue}), question answering (BoolQ \cite{clark2019boolq}), and common sense reasoning \cite{bisk2020piqa}). We also subsample large datasets to 4,000 samples. Therefore, PIQA has 3,674 samples, RTE has 552 samples, and the rest has 4,000 samples each.

\paragraph{License.} All datasets used in this work are publicly available and applied here only for non-commercial academic evaluation, consistent with the terms set by their original authors. Explicit licenses, where available, are: MIT \cite{de2019bias, zhao2018gender}, Apache 2.0 \cite{webster2018mind}, CC BY-SA 3.0 \cite{lehmann2015dbpedia, clark2019boolq}, Academic Free License v3.0 \cite{bisk2020piqa}, and CC BY 4.0 \cite{eight2016twitter, scheurer2024insider_trading}. The remaining datasets are released under research-use terms specified in their associated publications. All data is English-language.

\section{Additional empirical results}
\label{app:empirical}

\begin{figure*}[t]
\centering
\includegraphics[width=0.9\linewidth]{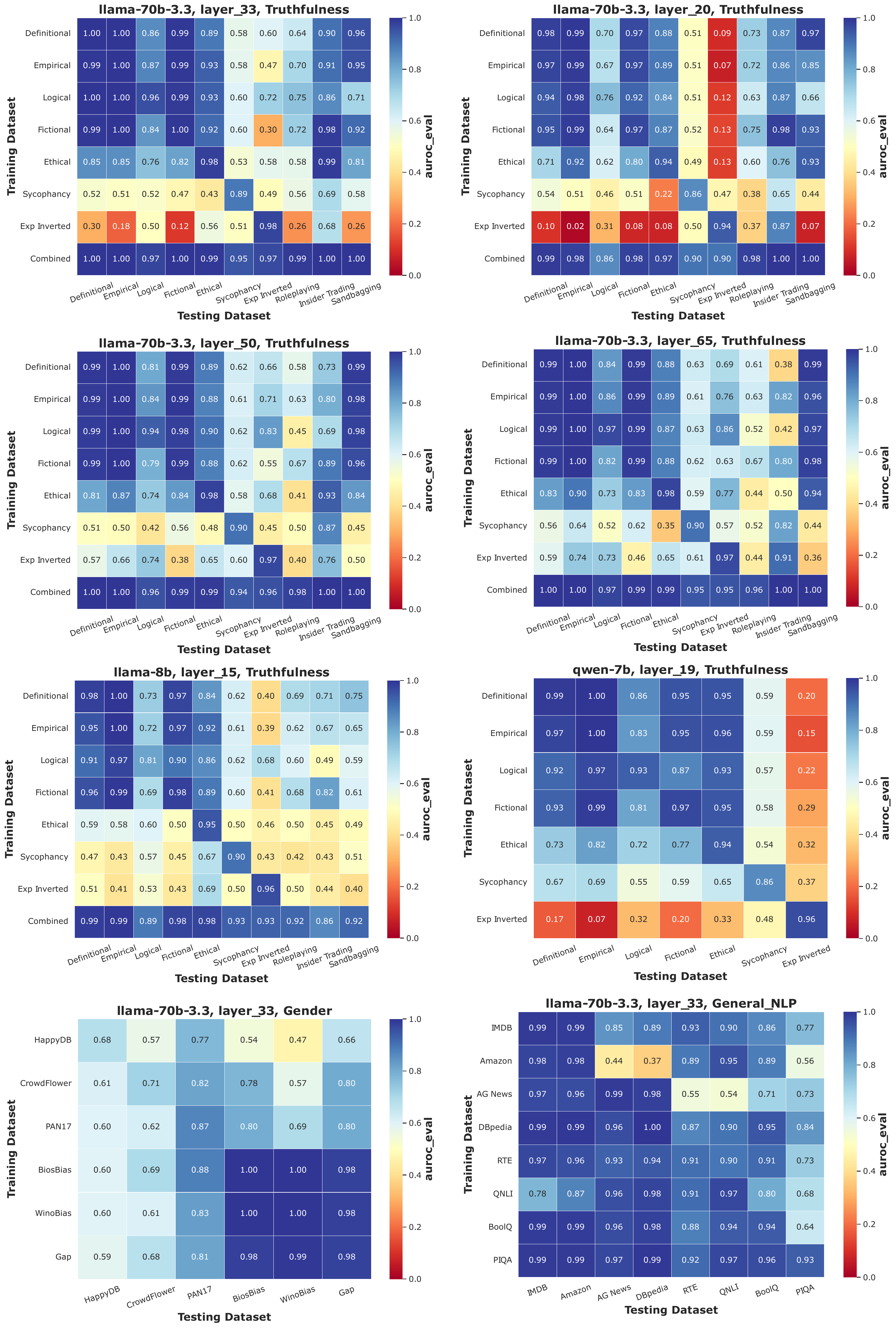}
\caption{\textbf{Cross-domain generalization performance for all eight conditions across models, layers, and concept domains.} We observe rich cross-domain generalization patterns across all eight conditions.}
\label{fig:full_auroc}
\end{figure*}

\paragraph{Cross-domain generalization performance.} As shown in Fig~\ref{fig:full_auroc}, we observe rich cross-domain generalization patterns across all eight conditions. 

\begin{figure*}[t]
\centering
\includegraphics[width=0.95\linewidth]{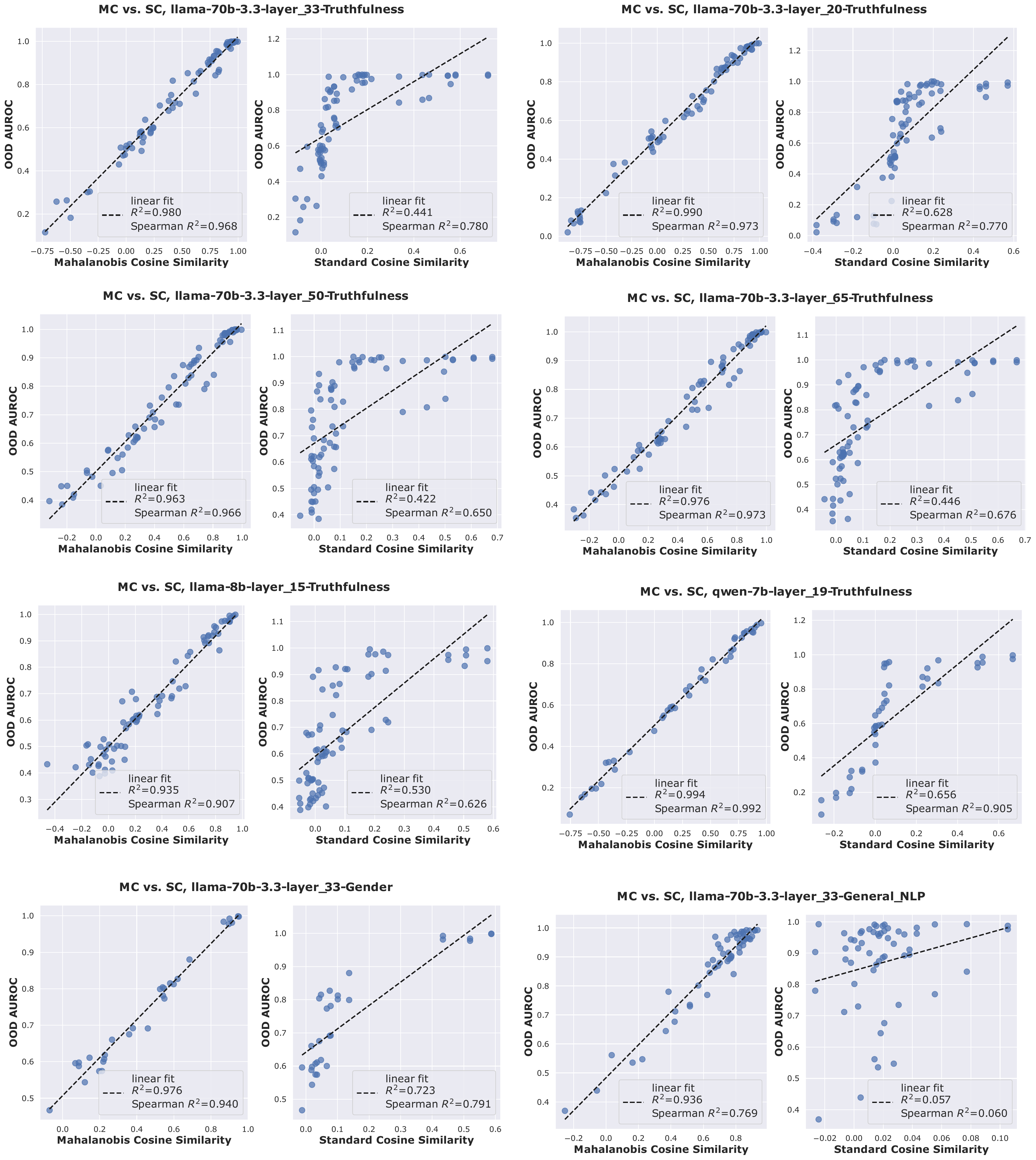}
\caption{\textbf{MCS and ECS against AUROC across conditions.} We observe a strong linear relationship between MCS and AUROC for all eight conditions across models, layers, and concept domains, while the relationship between ECS and AUROC is much weaker.}
\label{fig:full_mc_vs_sc}
\end{figure*}

\paragraph{MCS and ECS against AUROC.} Tab.~\ref{tab:r2} summarizes the performance of MCS and ECS on predicting OOD AUROC. We show the scatter plots for all eight conditions in Fig~\ref{fig:full_mc_vs_sc}. We observe right cross-domain generalization patterns across all eight conditions.  

\begin{figure*}[t]
\centering
\includegraphics[width=0.9\linewidth]{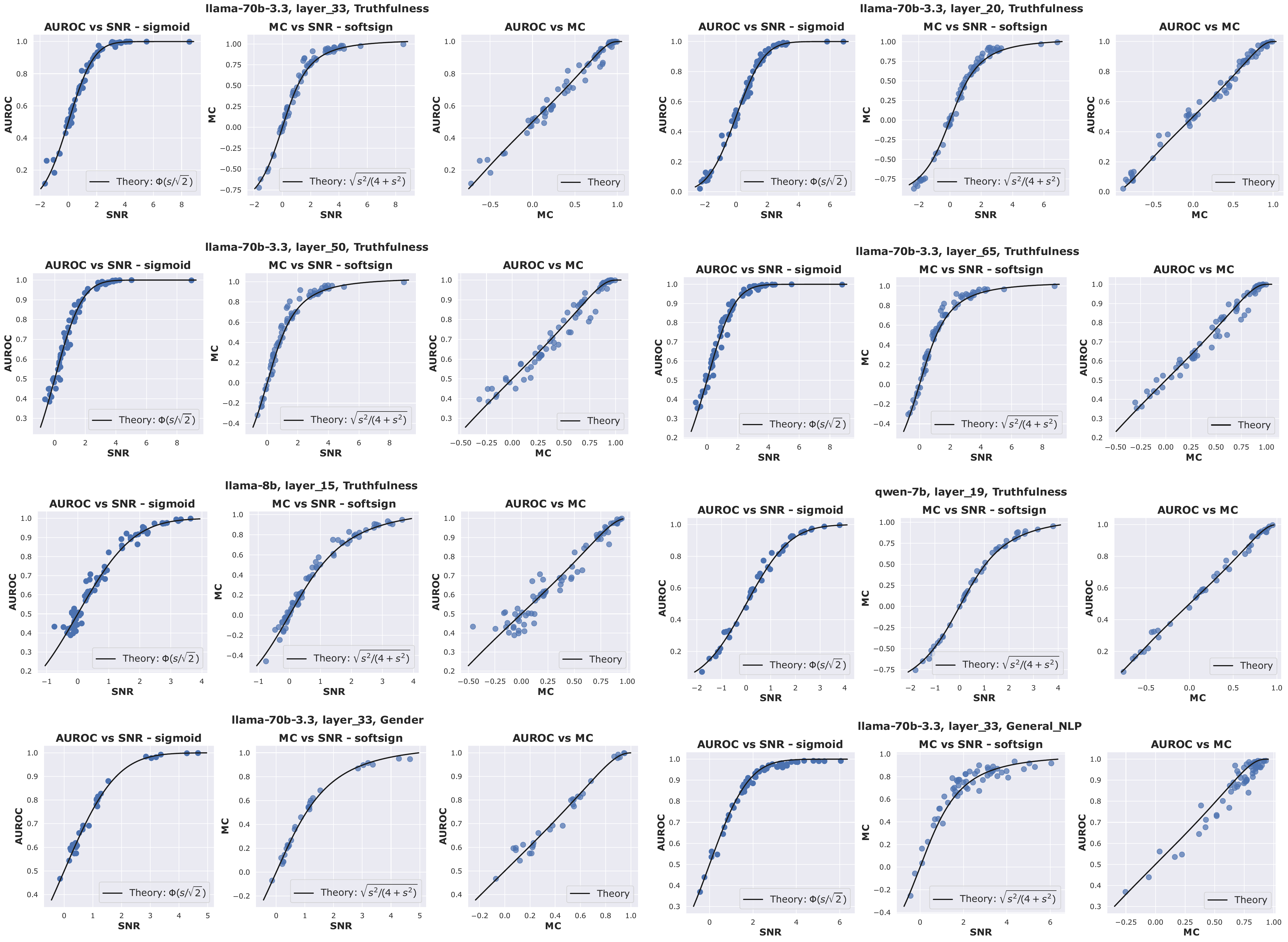}
\vspace{-8pt}
\caption{\textbf{Empirical verification of the theory.} The theory predicts the empirical data well across conditions.}
\label{fig:full_verify_theory}
\vspace{-15pt}
\end{figure*}

\paragraph{Verification of the theory across conditions.} We show the theory against empirical data across all eight conditions in Fig.~\ref{fig:full_verify_theory}. In all conditions, the theoretical prediction tracks the empirical data very well. 

\paragraph{Using $w_{ood}^{LDA}$ instead of $w_{ood}^{LR}$ for the empirical results.} We show the performance of using $\MC_{\Sigt}(\wid, \wood^{\mathrm{LR}})$ vs. $\MC_{\Sigt}(\wid, \wood^{\mathrm{LDA}})$ to predict the OOD generalization performance in Tab.~\ref{tab:r2_lda}. Using $\MC_{\Sigt}(\wid, \wood^{\mathrm{LDA}})$ is better than $\MC_{\Sigt}(\wid, \wood^{\mathrm{LR}})$ in all conditions, as expected. But the difference is small ($<0.05$).

\begin{table*}[t]
\centering\small
\begin{tabular}{lcccc}
\toprule
& \multicolumn{4}{c}{$R^2$} \\
\cmidrule(lr){2-5}
Condition & LR & LDA & LR-$\Sigma_{tot,diag}$ & LR-$\Sigma_{tot,LW}$ \\
\midrule
\texttt{Llama-70B} L33, truth & 0.980 & 0.987 & 0.438 & 0.980 \\
\midrule
\multicolumn{5}{l}{\emph{Layers} (\texttt{Llama-70B}, truth)} \\
\quad layer 20 & 0.990 & 0.990 & 0.620 & 0.990 \\
\quad layer 50 & 0.963 & 0.973 & 0.443 & 0.963 \\
\quad layer 65 & 0.976 & 0.978 & 0.422 & 0.975 \\
\midrule
\multicolumn{5}{l}{\emph{Domains} (\texttt{Llama-70B} L33)} \\
\quad Gender classification     & 0.976 & 0.983 & 0.675 & 0.977 \\
\quad General NLP benchmarks    & 0.936 & 0.982 & 0.002 & 0.936 \\
\midrule
\multicolumn{5}{l}{\emph{Models} (truth)} \\
\quad \texttt{Llama-3.1-8B}   & 0.935 & 0.962 & 0.438 & 0.934 \\
\quad \texttt{Qwen2.5-7B}     & 0.994 & 0.992 & 0.717 & 0.993 \\
\bottomrule
\end{tabular}
\caption{\textbf{Robustness check for the linear-fit of AUROC---MCS on MCS choices.} We probe two axes of robustness: (i)~the reference direction, comparing logistic regression (LR) $\MC_{\Sigt}(\wid, \wood^{\mathrm{LR}})$ against the theoretically motivated LDA direction $\MC_{\Sigt}(\wid, \wood^{\mathrm{LDA}})$, and the LDA version dominates the LR version, but the difference is small; (ii)~the shrinkage estimator for $\Sigma_{tot}$, comparing the full covariance against a per-coordinate diagonal approximation ($\MC_{tot,diag}(\wid, \wood^{\mathrm{LR}})$) and Ledoit--Wolf shrinkage ($\MC_{tot,LW}(\wid, \wood^{\mathrm{LR}})$). Ledoit--Wolf shrinkage is very similar to the full covariance we used, while the diagonal covariance is significantly worse.}
\label{tab:r2_lda}
\end{table*}

\begin{table*}[!htbp]
\centering
\small
\begin{tabular}{lccc}
\toprule
Condition & $\min z_{\max}$ & $\max z_{\max}$ & mean $z_{\max}$ \\
\midrule
Llama-70B L33, truth & 87.72 & 477.73 & 284.79 \\
\midrule
\multicolumn{4}{l}{\emph{Layers} (Llama-70B, truth)} \\
\quad layer 20 & 22.35 & 220.08 & 122.85 \\
\quad layer 50 & 90.53 & 925.78 & 533.96 \\
\quad layer 65 & 186.48 & 1377.71 & 852.16 \\
\midrule
\multicolumn{4}{l}{\emph{Domains} (Llama-70B L33)} \\
\quad Gender classification & 25.63 & 196.62 & 103.25 \\
\quad General NLP benchmarks & 41.50 & 379.83 & 142.83 \\
\midrule
\multicolumn{4}{l}{\emph{Models} (truth)} \\
\quad Llama-3.1-8B & 43.55 & 263.66 & 151.57 \\
\quad Qwen2.5-7B & 154.39 & 2724.74 & 1451.37 \\
\bottomrule
\end{tabular}
\caption{Empirical Fisher distance $\smash{ z_{\max} = \sqrt{\delta^\top \Sigma_{\text{pool}}^{-1} \delta}}$ per condition, computed on the OOD train half across all OOD tasks. All values have $z_{\max} \geq 20$, at which point the per-task slopes of AUROC--MCS curve lie within 0.5\% of their limits, making a single strong linear fit across all tasks possible (App.~\ref{app:slope}).}
\label{tab:zmax}
\end{table*}

\paragraph{Empirical AUROC--MCS slopes are consistent with theory.} As shown in Fig.~\ref{fig:slope}, the empirically fitted slopes across 8 conditions are consistent with the theoretical prediction from App.~\ref{app:slope}. 

\begin{figure*}[t]
\centering
\includegraphics[width=0.5\linewidth]{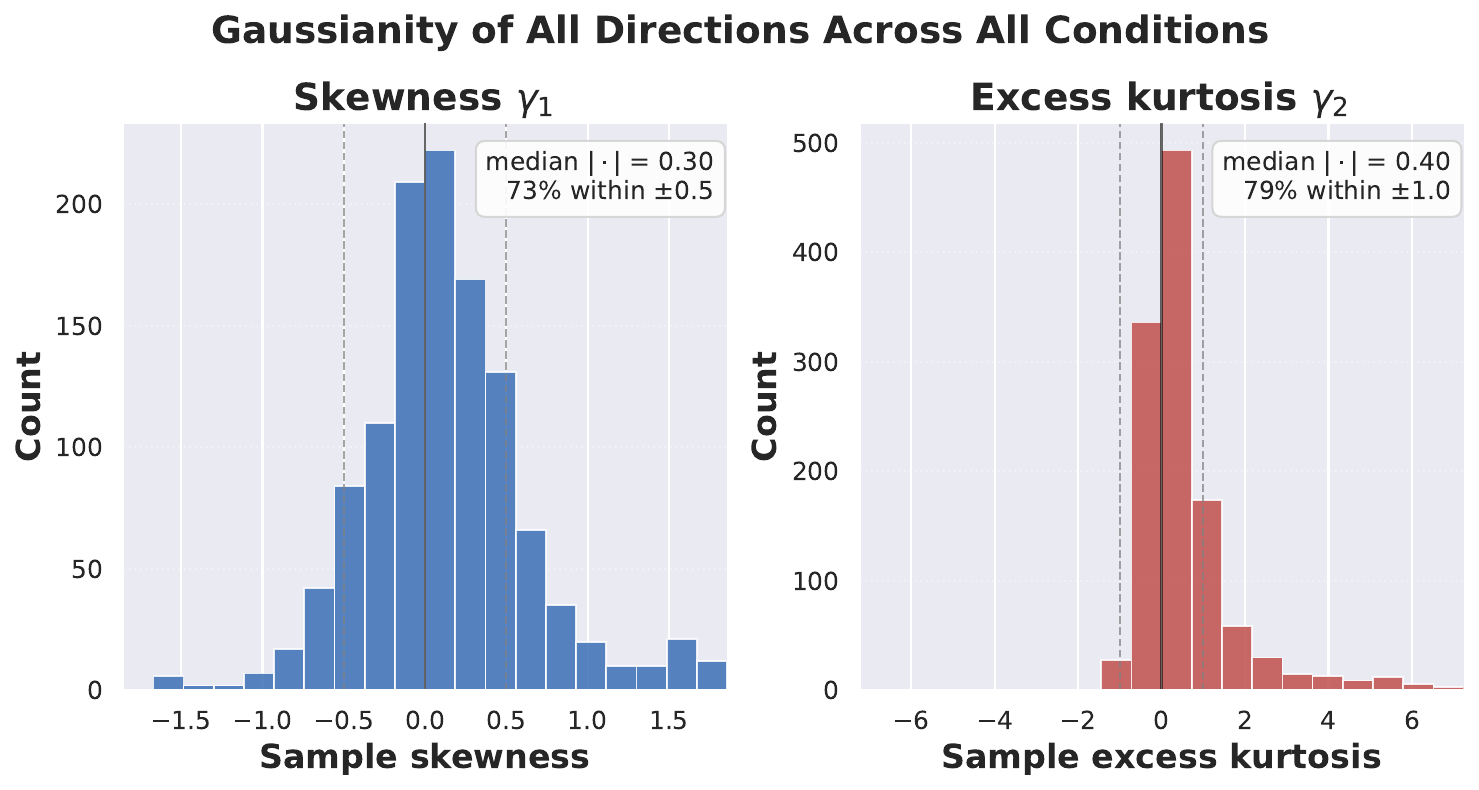}
\vspace{-8pt}
\caption{\textbf{Most directions are largely Gaussian across all conditions.} Across all conditions, all probe directions, and all test data distributions, most samples are largely Gaussian. Some samples have notably high kurtosis, which is mostly attributed to the sycophantic lying dataset.}
\label{fig:gaussian}
\vspace{-15pt}
\end{figure*}

\paragraph{Measuring Gaussianity of the empirical data.} As shown in Fig.~\ref{fig:gaussian}, most of the empirical skewness and kurtosis are small across all conditions, all probe directions, and all test data distributions. Specifically, 73\% of the skewness is within $\pm0.5$ and 79\% of the kurtosis is within $\pm 1$.

\begin{figure*}[t]
\centering
\includegraphics[width=0.9\linewidth]{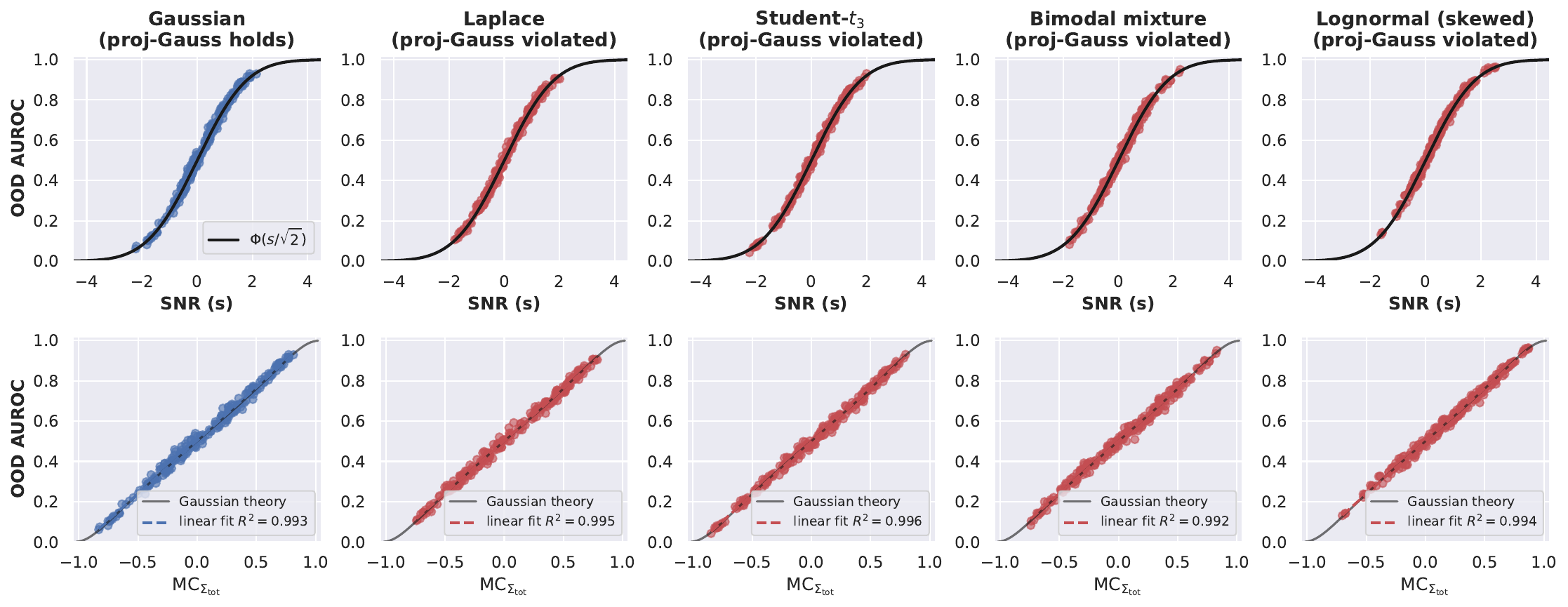}
\vspace{-8pt}
\caption{\textbf{The strong linearity between AUROC and MCS still holds for non-Gaussian distributions.} On deliberated constructed distributions where the projection-Gaussianity assumption is broken, the relationship between AUROC and MCS is still linear.}
\label{fig:non-gaussian}
\vspace{-15pt}
\end{figure*}

\paragraph{Simulating non-Gaussian distributions.} As shown in Fig.~\ref{fig:non-gaussian}, the AUROC--MCS $R^2$ is extremely high in simulations, even for distributions that are deliberately designed to violate the projection-Gaussianity assumption.

\paragraph{Empirical $\zmax$ across all datasets.} As shown in Tab.~\ref{tab:zmax}, all $\zmax$ across all conditions and datasets exceed the saturation threshold $z_{\max} \geq 20$ at which the slopes lie within 0.5\% of their limits (App.~\ref{app:slope}), pinning all per-task AUROC--MCS slopes to a universal slope. The average $\zmax$ for each condition is greater than 100, far exceeding the saturation point.

\paragraph{Robustness to the $\Sigt$ estimator.} Tab. \ref{tab:r2_lda} reports the AUROC–-MCS linear-fit $R^2$ under three estimators of $\Sigt$: the full sample covariance, Ledoit–Wolf shrinkage, and a per-coordinate diagonal approximation. Ledoit–Wolf is essentially indistinguishable from the full covariance across all conditions, so the headline result is not an artifact of a particular regularization choice. The diagonal approximation, by contrast, degrades sharply: most strikingly on general NLP ($R^2$=0.002 vs. 0.936), confirming that off-diagonal covariance structure is essential to MCS and that the linearity is not recoverable from per-coordinate variances alone.

\section{AI Usage}

We use AI assistants to assist with writing paper drafts, coding for experiments, and writing proofs. 

\clearpage

\section{Proofs of background results}
\label{app:classical-proofs}

\subsection{Proof of Lemma~\ref{lem:cov} (within--between covariance decomposition)}

By the law of total covariance,
\[
\Sigt = \E[\Cov(X \mid Y)] + \Cov(\E[X \mid Y]).
\]
The first term equals $\tfrac12 \Sigma_0 + \tfrac12 \Sigma_1 = \Sigp$. For the second, $\E[X \mid Y]$ takes the values $\mu_0$ and $\mu_1$ each with probability $\tfrac12$, and $\mu_c - \bar\mu = \pm\tfrac12 \delta$ (where $\bar\mu \coloneqq \tfrac12(\mu_0 + \mu_1)$), so
\[
\Cov(\E[X \mid Y]) = \tfrac12 \cdot \tfrac14 \delta\delta^\top + \tfrac12 \cdot \tfrac14 \delta\delta^\top = \tfrac14 \delta\delta^\top.
\]
\qed

\subsection{Proof of Lemma~\ref{lem:auroc} (binormal AUROC)}
By definition, $\AUROC(w) = \Pr(w^\top X_1 {>} w^\top X_0)$ where $X_c$ is drawn from the class-$c$ conditional. By Assumption~\ref{ass:gauss}, $w^\top X_c {\sim} \mathcal{N}(w^\top \mu_c, \sigma_c^2(w))$ with $\sigma_c^2(w) {\coloneqq} w^\top \Sigma_c w$, and $\sigma_0^2(w) + \sigma_1^2(w) = 2\,w^\top \Sigp w$. Given the standard assumption of the independence of $X_0, X_1$, we then have:

\begin{align*}
w^\top X_1 - w^\top X_0
  &\sim \mathcal{N}\bigl(w^\top \delta,\; \sigma_0^2(w) + \sigma_1^2(w)\bigr) \\
  &= \mathcal{N}\bigl(w^\top \delta,\; 2\,w^\top \Sigp w\bigr),
\end{align*}
hence
\begin{align*}
\AUROC(w) &= \Phi\bigl(w^\top \delta / \sqrt{2\,w^\top \Sigp w}\bigr) \\
&= \Phi(\SNR(w)/\sqrt 2).
\end{align*}
\qed

\subsection{Proof of Lemma~\ref{lem:fisher} (Fisher's discriminant)}

$\SNR^2(w) = (w^\top \delta)^2 / (w^\top \Sigp w)$ is scale-invariant, so we may restrict to $\{w : w^\top \Sigp w = 1\}$ and maximise $(w^\top \delta)^2$. Substituting $u \coloneqq \Sigp^{1/2} w$ and $q \coloneqq \Sigp^{-1/2} \delta$ converts the constraint to $\|u\| = 1$ and the objective to $(u^\top q)^2$. Cauchy--Schwarz gives a maximum of $\|q\|^2 = \delta^\top \Sigp^{-1} \delta = \zmax^2$, attained at $u \propto q$, i.e., $w \propto \Sigp^{-1/2} q = \Sigp^{-1} \delta = \wood$.
\qed

\section{Proof of Theorem~\ref{thm:mc}}
\label{app:thm-proof}

Write $v \coloneqq \wid^\top \Sigp \wid > 0$, so that $\wid^\top \delta = s\sqrt v$. Using Lemma~\ref{lem:cov} we have $\Sigt = \Sigp + \tfrac14 \delta\delta^\top$, and since $\wood = \Sigp^{-1}\delta$,
\[
\Sigp \wood = \delta, \delta^\top \wood = \delta^\top \Sigp^{-1} \delta = \zmax^2.
\]
We compute the three quadratic forms in~\eqref{eq:mc-def}.

\emph{Cross term.}
\begin{align*}
\wid^\top \Sigt \wood
&= \wid^\top \Sigp \wood + \tfrac14 \wid^\top \delta\, \delta^\top \wood \\
&= \wid^\top \delta + \tfrac14 (\wid^\top \delta)\,\zmax^2 \\
&= s\sqrt v\,(1 + \tfrac14 \zmax^2).
\end{align*}

\emph{Self-norm of $\wid$.}
\begin{align*}
    \wid^\top \Sigt \wid
&= v + \tfrac14 (\wid^\top \delta)^2 \\
&= v + \tfrac14 s^2 v \\
&= v(1 + \tfrac14 s^2).
\end{align*}

\emph{Self-norm of $\wood$.}
\[
\wood^\top \Sigt \wood
= \zmax^2 + \tfrac14 \zmax^4
= \zmax^2(1 + \tfrac14 \zmax^2).
\]

Substituting into~\eqref{eq:mc-def},
\begin{align*}
& \MC_{\Sigt}(\wid,\wood) \\
&= \frac{s\sqrt v\,(1 + \tfrac14 \zmax^2)}{\sqrt{v(1 + \tfrac14 s^2)} \cdot \sqrt{\zmax^2(1 + \tfrac14 \zmax^2)}} \\
&= \frac{s}{\zmax}\sqrt{\frac{1 + \tfrac14 \zmax^2}{1 + \tfrac14 s^2}}.
\end{align*}
Oddness, monotonicity, and the endpoint values $\MC_{\Sigt}(\pm\zmax) = \pm 1$ follow by inspection. Differentiating $s/\sqrt{1 + s^2/4}$ in $s$ yields
\begin{equation}
\frac{d\,\MC_{\Sigt}}{ds} = \frac{\sqrt{1 + \tfrac14 \zmax^2}}{\zmax} \cdot (1 + \tfrac14 s^2)^{-3/2},
\label{eq:mc-deriv}
\end{equation}
strictly positive on $(-\zmax, \zmax)$. \qed

\section{Slope along the AUROC--MCS curve}
\label{app:slope}
 
We derive the slope formula~\eqref{eq:slope-full} and prove the bound on the slope along the curve. All facts here follow from Lemma~\ref{lem:auroc} and Theorem~\ref{thm:mc} with no new assumptions.
 
\paragraph{Derivative of $\MC_{\Sigt}$ in $s$.}
Let $g(s) \coloneqq \MC_{\Sigt}(s) = (s/\zmax)\sqrt{(1+\alpha)/(1+s^2/4)}$ from~\eqref{eq:mc-tot}, with $\alpha = \zmax^2/4$. A direct calculation gives
\begin{equation}
g'(s) \;=\; \frac{\sqrt{1+\alpha}}{\zmax}\cdot\frac{1}{(1+s^2/4)^{3/2}} \;>\; 0,
\label{eq:gprime}
\end{equation}
so $g$ is strictly increasing on $(-\zmax,\zmax)$.
 
\paragraph{Slope formula~\eqref{eq:slope-full}.}
From Lemma~\ref{lem:auroc}, $\AUROC(s) = \Phi(s/\sqrt 2)$, so $d\AUROC/ds = \phi(s/\sqrt 2)/\sqrt 2$. Combining with~\eqref{eq:gprime},
\begin{align*}
\frac{d\AUROC}{d\MC_{\Sigt}}
&= \frac{d\AUROC/ds}{g'(s)} \\
&= \frac{\phi(s/\sqrt 2)/\sqrt 2}{(\sqrt{1+\alpha}/\zmax)(1+s^2/4)^{-3/2}},
\end{align*}
which simplifies to~\eqref{eq:slope-full}.
 
 

\paragraph{Factorization into task and shape factors.}
The slope~\eqref{eq:slope-full} factors as
\[
\frac{d\AUROC}{d\MC_{\Sigt}} \;=\; h(s)\cdot g(\zmax),
\]
where $h(s) = \phi(s/\sqrt 2)\,(1+s^2/4)^{3/2}$ depends only on $s$ (task-independent), and $g(\zmax) = 1/\sqrt{2/\zmax^2 + 1/2}$ depends only on $\zmax$ (independent of $s$). This decouples the two questions of (i) how flat the curve is within a task and (ii) how much the curve varies across tasks.

\paragraph{Task factor $g(\zmax)$ saturates fast.}
$g$ is strictly increasing in $\zmax$ with limit $g(\infty) = \sqrt 2$. Writing $g(\zmax)/g(\infty) = 1/\sqrt{1 + 4/\zmax^2}$, the deviation from the limit is at most $0.5\%$ for $\zmax \geq 20$ and falls off as $\zmax^{-2}$:
\[
\begin{array}{c|cccc}
\zmax        & 4 & 6 & 8 & 20 \\\hline
g(\zmax)/g(\infty) & 0.894 & 0.949 & 0.970 & 0.995
\end{array}
\]
Every empirical task has $\zmax > 20$ (Tab.~\ref{tab:zmax}), so each task's AUROC--MCS curve coincides with the $\zmax \to \infty$ limit curve to within $0.5\%$ on the slope.

\paragraph{Shape factor $h(s)$ and central slope.}
$h(s)$ combines $\phi(s/\sqrt 2)$ (strictly decreasing in $|s|$) with $(1+s^2/4)^{3/2}$ (strictly increasing in $|s|$). Their product is far flatter than either factor alone — this is the saturation cancellation behind (i): as $|s| \to \zmax$, the AUROC sigmoid saturates (shrinking $\phi$) at the same time as the MCS softsign saturates (growing
$(1+s^2/4)^{3/2}$), and the two effects largely offset. At the centre, $h(0) = 1/\sqrt{2\pi}$, giving the universal central slope
\[
h(0)\cdot g(\infty) \;=\; \frac{\sqrt 2}{\sqrt{2\pi}} \;=\; \frac{1}{\sqrt\pi}
\;\approx\; 0.564.
\]
The cancellation is not exact: $h(s)\cdot g(\infty)$ peaks at $s=\pm\sqrt 2$ with value ${\approx}0.629$, then decays toward $0$ as $|s| \to \zmax$, which is why $R^2 < 1$ in the empirical AUROC--MCS fits.

\begin{figure}[t]
\centering
\includegraphics[width=0.95\linewidth]{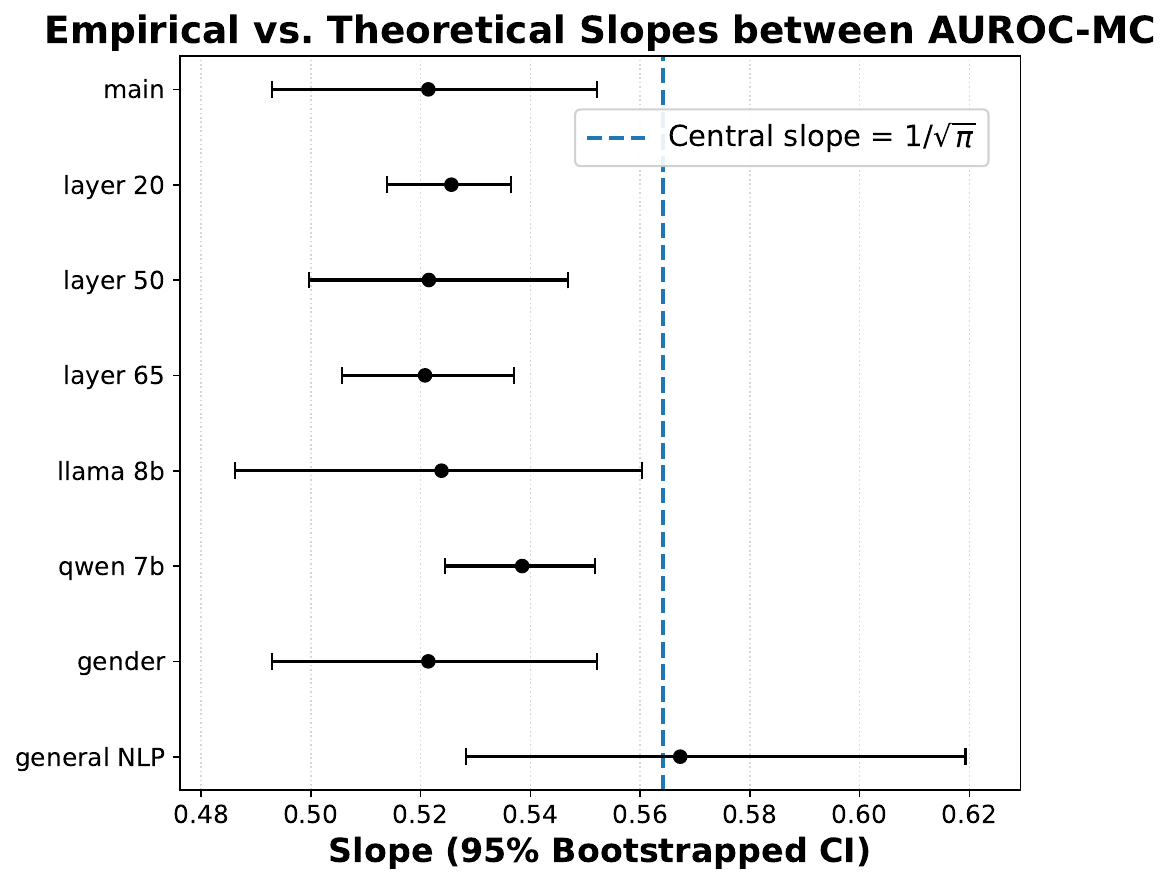}
\caption{\textbf{Empirical slopes are near the theoretical central slope.} Linear-fit slope of OOD AUROC against $\MC_{\Sigt}$, with 95\% bootstrap CIs, across all eight conditions. The dashed line marks the universal central slope $1/\sqrt\pi \approx 0.564$ predicted by the theory (App.~\ref{app:slope}). Empirical estimates are close to this value, with most of them slightly below it. This is consistent with data sampling some of the saturation tail where the local slope decays, dragging down the overall slope.}
\label{fig:slope}
\end{figure}

\paragraph{Empirical slopes are consistent with theory.} As shown in Fig.~\ref{fig:slope}, the empirically fitted slopes across 8 conditions are consistent with the theoretical prediction. 

\section{Failure mode details}
\label{app:failures}

This appendix collects the formal statements behind the failure modes summarised in \S\ref{sec:failures}.

\subsection{Use \texorpdfstring{$\Sigp$}{Sigma\_pool} metric instead of \texorpdfstring{$\Sigt$}{Sigma\_tot}}

We have: 
\[
\wid^\top \Sigp \wood = \wid^\top \delta = s\sqrt v;\ \ \wid^\top \Sigp \wid = v
\]
and 
$$\wood^\top \Sigp \wood = \delta^\top \Sigp^{-1} \delta = \zmax^2$$
Substituting into~\eqref{eq:mc-def} gives $s\sqrt v / (\sqrt v \cdot \zmax) = s/\zmax$. Therefore, $\MC_{\Sigp}$ is just a rescaled $s$. This will not cancel out the S-shaped between OOD AUROC and $s$, and thus will not be linear to OOD AUROC.

\qed

\subsection{Small Fisher distance}

When $\zmax$ is small ($\zmax \lesssim 2$), the slope around 0 $\kappa(\zmax)$ is far below its limit and varies rapidly with $\zmax$: $\kappa(1) \approx 0.25$ vs.\ $\kappa(2) \approx 0.40$ vs.\ $\kappa(3) \approx 0.47$. A collection of test sets with heterogeneous small-$\zmax$ tasks therefore does not collapse onto a single linear fit; each task has its own slope, producing a fan rather than a line. On synthetic data with $\zmax$ controlled to $\{0.1, 0.5, 1, 2\}$ (Fig.~\ref{fig:failures}c), we observe exactly this fan, and the global linear $R^2$ drops to 0.666. This failure mode is not engaged on our LLM datasets because every task has $\zmax > 20$, inside the saturation regime (see Tab.~\ref{tab:zmax}; App.~\ref{app:empirical}).

\subsection{Strong class imbalance}

For class prior $\pi$, we redefine the pooled vvariance as:
$$\Sigp \coloneqq (1-\pi)\Sigma_0 + \pi \Sigma_1$$
Then Lemma~\ref{lem:cov} generalises to 
$$\Sigt = \Sigp + \pi(1-\pi)\,\delta\delta^\top$$ , and Theorem~\ref{thm:mc} becomes
\[
\MC_{\Sigt}(\wid, \wood) {=} \frac{s}{\zmax}\sqrt{\frac{1 + \pi(1-\pi)\zmax^2}{1 + \pi(1-\pi)s^2}},
\]
preserving the softsign-cancels-sigmoid story but with a re-scaled saturation point and slope. Strong imbalance (e.g., $\pi = 0.01$) shrinks the $\pi(1-\pi)$ factor by $25\times$ versus the balanced case, weakening the saturation cancellation in MCS and steepening the AUROC--MCS slope. Our probe datasets are all approximately balanced, so this mode is not engaged in our experiments, but it would matter for naturally imbalanced applications such as rare-event detection.

\section{Empirical alignment of LR and LDA reference directions}
\label{app:lr-lda}

The theory in \S\ref{sec:theory} uses the Fisher direction $\wood^{\mathrm{LDA}} = \Sigp^{-1}\delta$ as the OOD reference. In the empirical experiments, we replace it with the OOD-trained logistic-regression direction $\wood^{\mathrm{LR}}$, on the grounds that for balanced binary tasks with well-separated classes the two directions are nearly proportional. This appendix quantifies the substitution.

\paragraph{Setup.} For each of datasets used in \S\ref{sec:empirical}, on the train half of a stratified 50/50 split, we compute (i) the Fisher direction $\wood^{\mathrm{LDA}} = (\Sigp + \lambda I)^{-1}\delta$ with $\lambda = 10^{-6}$, and (ii) the LR direction $\wood^{\mathrm{LR}}$. We measure the correlation between MC($w^{LR}_{id}$, $w^{LR}_{ood}$) and MC($w^{LR}_{id}$, $w^{LDA}_{ood}$).

\begin{figure}[t]
\centering
\includegraphics[width=0.98\linewidth]{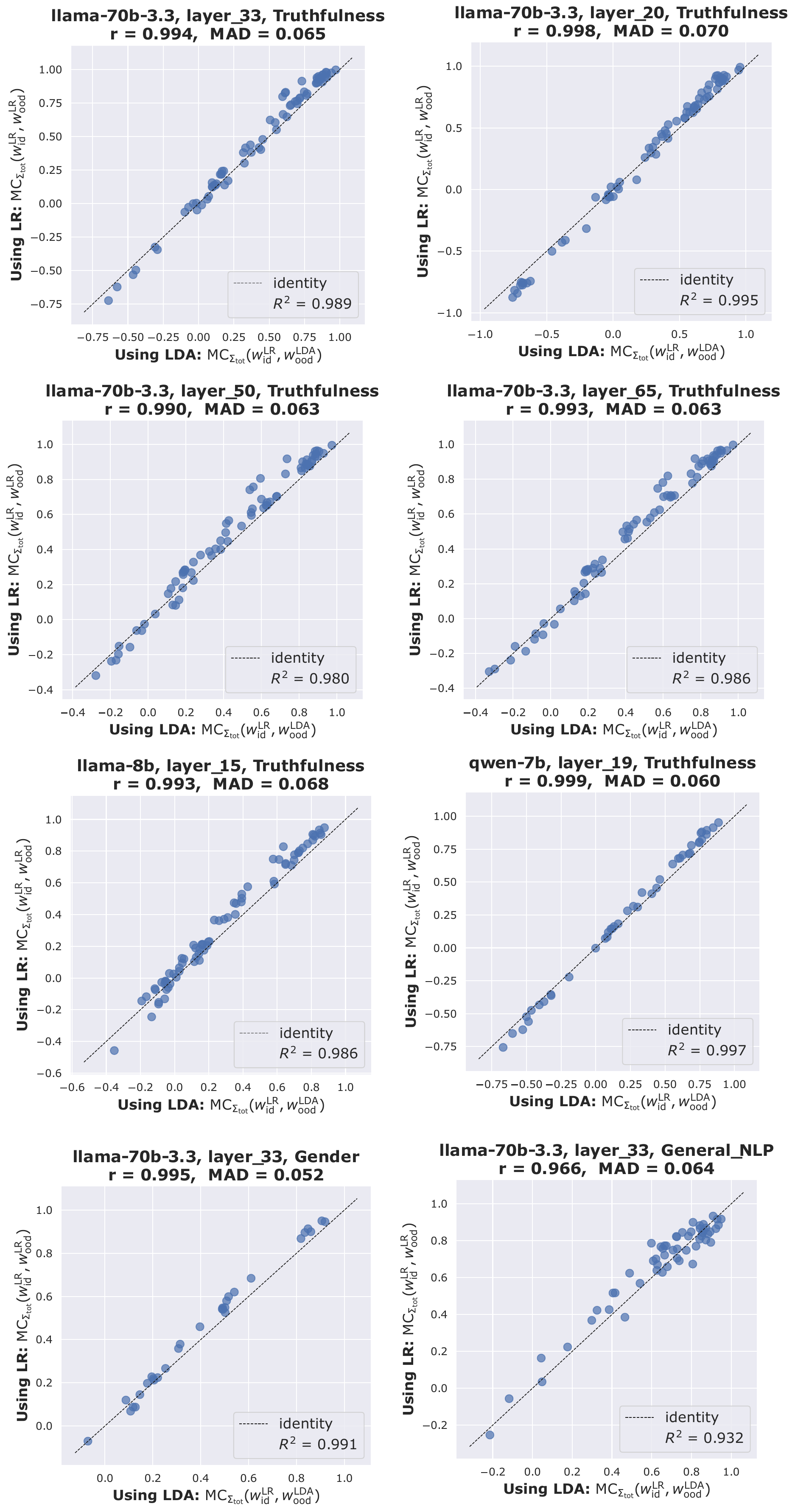}
\vspace{-8pt}
\caption{\textbf{MC($w^{LR}_{id}$, $w^{LR}_{ood}$) vs. MC($w^{LR}_{id}$, $w^{LDA}_{ood}$).} The two MCs correlate strongly across all eight conditions, explaining why substituting the LDA direction with the LR direction in our empirical experiments does not affect the observed linearity.}
\label{fig:lr_vs_lda}
\vspace{-15pt}
\end{figure}

\paragraph{Result.} As shown in Fig.~\ref{fig:lr_vs_lda}, across all settings (across models, layers, and concept domains), the pairwise Pearson correlation between $\MC_{\Sigt}(\wid, \wood^{\mathrm{LR}})$ and $\MC_{\Sigt}(\wid, \wood^{\mathrm{LDA}})$ is $r > 0.99$, with mean $|\Delta(\wid)| < 0.07$. Substituting $\wood^{\mathrm{LR}}$ for $\wood^{\mathrm{LDA}}$ in the headline regression of \S\ref{sec:empirical} changes the linear-fit $R^2$ by less than 1\%.

\paragraph{Why this works.} For balanced binary classification with classes that are jointly well-modelled by Gaussians with shared covariance, the LR maximum-likelihood direction and the Fisher direction coincide up to scaling \cite{efron1975efficiency, hastie2009elements}. Real activations only approximately satisfy these assumptions, but all of our probe datasets are easy --- ID AUROC exceeds $0.95$ in every case --- so the residual mismatch between $\wood^{\mathrm{LR}}$ and $\wood^{\mathrm{LDA}}$ is small in cosine. We expect a larger mismatch on harder, less-separable tasks; characterising that regime is a useful direction for future work.

\end{document}